\documentclass[runningheads, envcountsame, a4paper]{llncs}
\usepackage[pdftex]{graphicx}
\usepackage{epstopdf}
\usepackage{calc}
\usepackage[hyphens]{url}
\usepackage{xcolor}
\usepackage{paralist}
\usepackage{amssymb}
\usepackage{amsmath}
\usepackage{wrapfig}
\usepackage{subfigure}
\usepackage[abs]{overpic}
\usepackage{calc}
\usepackage[misc]{ifsym}

\begin{document}

\title{Meta-Learning for Black-box Optimization}
\titlerunning{Meta-Learning for Black-box Optimization}
\toctitle{Meta-Learning for Black-box Optimization}
%
\author{Vishnu TV (\Letter) \and
Pankaj Malhotra\and
Jyoti Narwariya\and
Lovekesh Vig\and \\
Gautam Shroff}
\authorrunning{V. TV et al.}

\tocauthor{Vishnu TV,Pankaj Malhotra,Jyoti Narwariya,Lovekesh Vig,Gautam Shroff}
%
\institute{TCS Research, New Delhi, India\\
\email{$\{$vishnu.tv,malhotra.pankaj,jyoti.narwariya,\\lovekesh.vig,gautam.shroff$\}$@tcs.com}\\
}
\maketitle              
\begin{abstract}
Recently, neural networks trained as optimizers under the ``learning to learn" or meta-learning framework have been shown to be effective for a broad range of optimization tasks including derivative-free black-box function optimization. 
Recurrent neural networks (RNNs) trained to optimize a diverse set of synthetic non-convex differentiable functions via gradient descent have been effective at optimizing derivative-free black-box functions.
In this work, we propose \textit{RNN-Opt}: an approach for learning RNN-based optimizers for optimizing real-parameter single-objective continuous functions under limited budget constraints.
Existing approaches utilize an observed improvement based meta-learning loss function for training such models. 
We propose training RNN-Opt by using synthetic non-convex functions with known (approximate) optimal values by directly using discounted regret as our meta-learning loss function. 
We hypothesize that a regret-based loss function mimics typical testing scenarios, and would therefore lead to better optimizers compared to optimizers trained only to propose queries that improve over previous queries.
Further, RNN-Opt incorporates simple yet effective enhancements during training and inference procedures to deal with the following practical challenges: 
i) Unknown range of possible values for the black-box function to be optimized, and 
ii) Practical and domain-knowledge based constraints on the input parameters.
We demonstrate the efficacy of RNN-Opt in comparison to existing methods on several synthetic as well as standard benchmark black-box functions along with an anonymized industrial constrained optimization problem.

\end{abstract}

\section{Introduction}
Several practical optimization problems such as process black-box optimization for complex dynamical systems pose a unique challenge owing to the restriction on the number of possible function evaluations. Such black-box functions do not have a simple closed form but can be evaluated (queried) at any arbitrary query point in the domain.  However, evaluation of real-world complex processes is expensive and time consuming, therefore the optimization algorithm must  optimize while employing as few real-world function evaluations as possible. Most practical optimization problems are constrained in nature, i.e. have one or more constraints on the values of input parameters. In this work, we focus on real-parameter single-objective black-box optimization (BBO) where the goal is to obtain a value as close  to the maximum value of the objective function as possible by adjusting the values of  the real-valued continuous input parameters while ensuring domain constraints are not violated. We further assume a limited budget, i.e. assume that querying the black-box function is expensive and thus only a small number of queries can be made. 

Efficient global optimization of expensive black-box functions \cite{jones1998efficient} requires proposing the next query (input parameter values) to the black-box function based on past queries and the corresponding responses (function evaluations).
BBO can be mapped to the problem of proposing the next query given past queries and the corresponding responses such that the expected improvement in the function value is maximized, as in Bayesian Optimization approaches \cite{brochu2010tutorial}. 
While most research in optimization has focused on engineering algorithms catering to specific classes of problems, recent meta-learning \cite{schmidhuber1987evolutionary} approaches, e.g. \cite{andrychowicz2016learning,li2016learning,chen2017learning,wichrowska2017learned,faury2018rover}, cast design of an optimization algorithm as a learning problem rather than the traditional hand-engineering approach, and then, propose approaches to train neural networks that \textit{learn to optimize}.
In contrast to a traditional machine learning approach involving training of a neural network on a single task using training data samples so that it can generalize to unseen data samples from the same data distribution, here the neural network is trained on a distribution of similar tasks (in our case optimization tasks) so as to learn a strategy that generalizes to related but unseen tasks from a similar task distribution.
The meta-learning approaches attempt to train a single network to optimize several functions at once such that the network can effectively generalize to optimize unseen functions.

Recently, \cite{chen2017learning} proposed a meta-learning approach wherein a recurrent neural network (RNN with gated units such as Long Short Term Memory (LSTM) \cite{hochreiter1997long}) learns to optimize a large number of diverse synthetic non-convex functions to yield a learned task-independent optimizer. The RNN iteratively uses the sequence of past queries and corresponding responses to propose the next query in order to maximize the observed improvement (OI) in the response value. We refer to this approach as RNN-OI in this work.
Once the RNN is trained to optimize a diverse set of synthetic functions by using gradient descent, it is able to generalize well to solve unseen derivative-free black-box optimization problems \cite{chen2017learning,zhou2017optimizing}.
Such learned optimizers are shown to be faster in terms of the time taken to propose the next query compared to Bayesian optimizers as they do not require any matrix inversion or optimization of acquisition functions, and also have lower regret values within the training horizon, i.e. the number of steps of the optimization process for which the RNN is trained to generate queries.

Key contributions of this work and the challenges addressed can be summarized as follows:
\begin{enumerate}
	\item \textit{Regret-based loss function}: We hypothesize that training an RNN optimizer using a loss function that minimizes the regret observed for a given number of queries more closely resembles the performance measure of an optimizer. So it is better than a loss function based on OI such as the one used in \cite{chen2017learning,zhou2017optimizing}. To this end, we propose a simple yet highly effective loss function that yields superior results than the existing OI loss for black-box optimization. Regret of the optimizer is the difference between the optimal value (maximum of the black-box function) and the realized maximum value. 
	\item \textit{Deal with lack of prior knowledge on range of the black-box function}: In many practical optimization problems, it may be difficult to ascertain the possible range of values the function can take, and the range of values would vary across applications. On the other hand, neural networks are known to work well only on normalized inputs, and can be numerically unstable and difficult to train on very large or very small values as typical non-linear activation functions like sigmoid activation function tend to saturate for large inputs and will then adjust slowly during training. 
	RNNs are most easily trained when their inputs are well conditioned, and have a similar scale as their latent state, and suitable scaling often accelerates training \cite{wichrowska2017learned}.
	We, therefore, propose \textit{incremental normalization} that dynamically normalizes the output (response) from the black-box function using the response values observed so far before the value is passed as an input to the RNN, and observe significant improvements in terms of regret by doing so.
	\item \textit{Incorporate domain-constraints}: Any practical optimization problem has a set of constraints on the input parameters. It is important that the RNN optimizer is penalized when it proposes query points outside the desired limits. We introduce a mechanism to achieve this by giving an additional feedback to the RNN whenever it proposes a query that violates domain constraints. In addition to regret-based loss, RNN is also trained to simultaneously minimize domain constraint violations. We show that an RNN optimizer trained in this manner attains lower regret values in fewer steps when subjected to domain constraints compared to an RNN optimizer not explicitly trained to utilize feedback.
\end{enumerate}
We refer to the proposed approach as \textit{RNN-Opt}. 
As a result of the above considerations, RNN-Opt can deal with an unknown range of function values and also incorporate domain constraints.
We demonstrate that RNN-Opt works well on optimizing unseen benchmark black-box functions and outperforms RNN-OI in terms of the optimal value attained under a limited budget for 2-dimensional and 6-dimensional input spaces. We also perform extensive ablation experiments demonstrating the importance of each of the above-stated features in RNN-Opt.

The rest of the paper is organized as follows: We contrast our work to existing literature in Section \ref{sec:rw}, followed by defining the problem in Section \ref{sec:overview}. We present the details of our approach in Section \ref{sec:rnn-opt}, followed by experimental evaluation in Section \ref{sec:exp}, and conclude in Section \ref{sec:discussion}.

\section{Related Work\label{sec:rw}}
Our work falls under the category of real-parameter black-box global optimization \cite{rios2013derivative}.
Traditional approaches for black-box optimization like covariance matrix adaptation evolution strategy (CMA-ES) \cite{hansen1996adapting}, Nelder-Mead \cite{nelder1965simplex}, and Particle Swarm Optimization (PSO) \cite{kennedy2010particle} hand-design rules using heuristics (e.g. using nature-inspired genetic algorithms) to decide the next query point(s) given the observations made so far.
Another category of approaches for global optimization of black-box functions include Bayesian optimization techniques \cite{brochu2010tutorial,snoek2012practical,shahriari2016taking}. These approaches use observations (query and response) made thus far to approximate the black-box function via a surrogate (meta-) model, e.g. using a Gaussian Process \cite{huang2006global}, and then use this model to construct an acquisition function to decide the next query point.  The acquisition function updates needed at each step are known to be costly \cite{chen2017learning}.

\textit{Learned optimizers}: There has been a recent interest in learning optimizers under the meta-learning setting \cite{schmidhuber1987evolutionary} by
training RNN optimizers via gradient descent.
For example, \cite{andrychowicz2016learning} casts the design of an optimization algorithm as a learning problem and uses an LSTM model to learn an optimizer for a particular class of optimization problems, e.g. quadratic functions, training neural networks, etc.
Similarly, \cite{li2016learning,faury2018rover} cast optimizer learning as learning a policy under a reinforcement learning setting. 
\cite{wichrowska2017learned} proposes a hierarchical RNN architecture to learn optimizers that scale well to optimize a large number of parameters (high-dimensional input space). 
However, the above meta-learning approaches for optimization assume the availability of gradient information to decide the next set of parameters, which is not available in the case of black-box optimization. 
Our work builds upon the meta-learning approach for learning black-box optimizers proposed in \cite{chen2017learning}. This approach mimics the sequential model-based Bayesian approaches in the sense that it proposes an RNN optimizer that stores sequential information about previous queries and responses, and accesses this memory to generate the next candidate query.
RNN-OI mimics the Bayesian optimization based sequential decision-making process \cite{brochu2010tutorial} (refer \cite{chen2017learning} for details) while being significantly faster than standard BBO algorithms like SMAC \cite{hutter2011sequential} and Spearmint \cite{snoek2012practical} as it does not involve any matrix inversion or optimization of acquisition functions. RNN-OI was successfully 
tested on Gaussian process bandits, simple low dimensional controllers, and hyper-parameter tuning.

\textit{Handling domain constraints in neural networks}
Recent work on Physics-guided deep learning \cite{jia2018physics,muralidhar2018incorporating} incorporates domain knowledge in the learning process via additional loss terms. Such approaches can be useful in our setting if the optimizer network is to be trained from scratch for a given application. However, the purpose of building a generic optimizer that can be transferred to new applications requires incorporating domain constraints in a posterior manner during inference time when the optimizer is suggesting query points. This is not only useful to adapt the same optimizer to a new application but also useful in another practical scenario of adapting to a new set of domain constraints for a given application.
ThermalNet \cite{cheng2018thermalnet} uses a deep Q-network as an optimizer and uses an LSTM predictor for combustion optimization of a boiler in a power plant but does not handle domain constraints. 
Similar to our approach, ChemOpt \cite{zhou2017optimizing} uses an RNN based optimizer for chemical reaction optimization but does not address aspects related to handling an unknown range for the function being optimized and incorporating domain constraints.

\textit{Handling unknown range of function values}:
Suitable scaling of input and output of hidden layers in neural networks has been shown to accelerate training of neural networks \cite{ioffe2015batch,salimans2016weight,ba2016layer,klambauer2017self}. Dynamic input scaling has been used in a similar setting as ours \cite{wichrowska2017learned} to ensure that the neural network based optimizer is invariant to parameter scale. 
However, the scaling is applied to the average gradients. In our setting, we use a similar approach but apply dynamic scaling to the function evaluations being fed back as input to RNN-Opt.

\section{Problem Overview\label{sec:overview}}
We consider learning an optimizer that can optimize (e.g., maximize) a black-box function $f_b: \Theta \mapsto \mathbb{R}$, where $\Theta \subseteq \mathbb{R}^d$ is the domain of the input parameters. We assume that the function $f_b$ does not have a closed-form representation, is costly to evaluate, and does not allow the computation of gradients. 
In other words, the optimizer can query the function $f_b$ at a point $\mathbf{x}$ to obtain a response $y=f_b(\mathbf{x})$, but it does not obtain any gradient information, and in particular it cannot make any assumptions on the analytical form of $f_b$.
The goal is to find $\mathbf{x}_{opt} = \mathrm{arg} \max_{\mathbf{x}\in \Theta} f_b(\mathbf{x})$ within a limited budget, i.e. within a limited number of queries $T$ that can be made to the black-box. 

We consider training an optimizer $f_{opt}$ with parameters $\boldsymbol{\theta}_{opt}$ such that, given the queries $\mathbf{x}_{1\ldots t} = \mathbf{x}_1,\mathbf{x}_2,\ldots, \mathbf{x}_{t}$ and the corresponding responses $y_{1\ldots t}= y_1,y_2,\ldots, y_{t}$ from $f_b$ where $y_t = f_b(\mathbf{x}_t)$, $f_{opt}$ proposes the next query point $\mathbf{x}_{t+1}$ under a budget constraint of $T$ queries, i.e. $t\leq T-1$:
\begin{align}
\mathbf{x}_{t+1}  &= f_{opt}(\mathbf{x}_{1\ldots t}, y_{1\ldots t};\boldsymbol{\theta}_{opt}).
\end{align}

\section{RNN-Opt \label{sec:rnn-opt}}
We model $f_{opt}$ using an LSTM-based RNN.
(For implementation, we use a variant of LSTMs as described in \cite{zaremba2014recurrent}.)
Recurrent Neural Networks (RNNs) with gated units such as Long Short Term Memory (LSTM) \cite{hochreiter1997long} units are a popular choice for sequence modeling to make predictions about future values given the past.
They do so by maintaining a memory of all the relevant information from the sequence of inputs observed so far.
In the meta-learning or training phase, a diverse set of synthetically-generated differentiable non-convex functions (refer Appendix \ref{apx:gmm-df}) with known global optima are used to train the RNN (using gradient descent). The RNN is then used to predict the next query in order to intelligently explore the search space given the sequence of previous queries and the function responses. 
The RNN is expected to learn to retain any information about previous queries and  responses that is relevant to proposing the next query to minimize the regret as shown in Fig. \ref{fig:RNN-Opt}.

\subsection{RNN-Opt without Domain Constraints}
Given a trained RNN-based optimizer and a differentiable function $f_g$, inference in RNN-Opt follows the following iterative process for $t=1,\ldots,T-1$:
At each step $t$, the output of the final recurrent hidden layer of the RNN is used to generate the output via an affine transformation to finally obtain $\mathbf{x}_{t+1}$.
\begin{align}
	\mathbf{h}_{t+1} &= f_o(\mathbf{h}_{t}, \mathbf{x}_{t}, y_{t};\boldsymbol{\theta}),\label{eq:rnn-opt}\\
	\boldsymbol{\mu}_{t+1}^x, \mathbf{\Sigma}_{t+1}^x &= W_{2m,d}(\mathbf{h}_{t+1}),\label{eq:mu-sig}\\
	\mathbf{x}_{t+1} & \sim \mathcal{N}(\boldsymbol{\mu}_{t+1}^x, \mathbf{\Sigma}_{t+1}^x),\label{eq:gen_x}\\
	y_{t+1} &=f_g(\mathbf{x}_{t+1}),
\end{align}
where $f_o$ represents the RNN with parameters $\boldsymbol{\theta}$, $f_g$ is the function to be optimized, $W_{2m,d}$ defines the affine transformation of the final output (hidden state) $\mathbf{h}_{t+1}$ of the RNN. The parameters $\boldsymbol{\theta}$ and $W_{2m,d}$ together constitute $\boldsymbol{\theta}_{opt}$. Instead of directly training $f_o$ to propose the next query $\mathbf{x}_{t+1}$ as in \cite{chen2017learning}, we use a stochastic RNN to estimate $\boldsymbol{\mu}_{t+1}^x \in \mathbb{R}^d$ and $\mathbf{\Sigma}_{t+1}^x \in \mathbb{R}^{d\times d}$ as in Equation \ref{eq:mu-sig}, then sample $\mathbf{x}_{t+1}$ from a multivariate Gaussian distribution $\mathcal{N}(\boldsymbol{\mu}_{t+1}^x, \mathbf{\Sigma}_{t+1}^x)$. Introducing randomness in the query generation process leads to better exploration compared to a deterministic model \cite{zhou2017optimizing}. The first query $\mathbf{x}_{1}$ is sampled from a uniform distribution over the domain of the function $f_g$ to be optimized. Once the network is trained, $f_g$ can be replaced by any black-box function $f_b$ that takes $d$-dimensional input.

For any synthetically generated function $f_g \in \mathcal{F}$, we assume $\mathbf{x}_{opt}$ (approximate) can be found, e.g. using gradient-descent, since the closed form of the function is known. Hence, we assume that $y_{opt}$ of $f_g$ given by $y_{opt}=f_g(\mathbf{x}_{opt})$ is known. 
Therefore, it is easy to determine the regret $y_{opt} - \max_{i\leq t} y_i$ after $t$ iterations (queries) to the function $f_g$.
We can then define a regret-based loss function as follows:
\begin{figure}
	\centering
	\includegraphics[scale=0.38,trim={0cm, 4cm, 0cm, 2.5cm},clip]{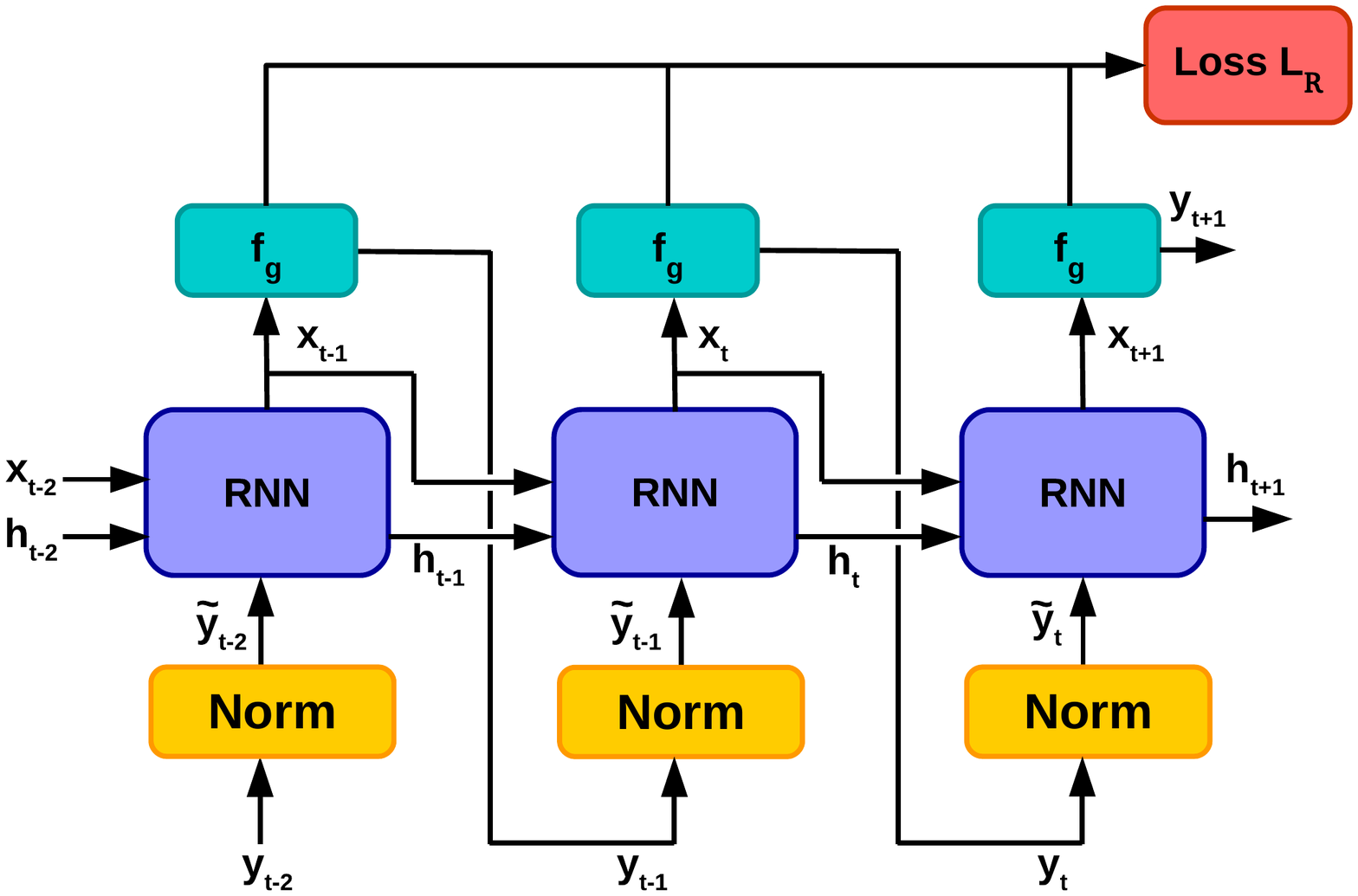}
	\caption{Computation flow in RNN-Opt. During training, the functions $f_g$ are differentiable and obtained using Equation \ref{eq:gmm}. Once trained, $f_g$ is replaced by the black-box function $f_b$.\label{fig:RNN-Opt}}
\end{figure}

\begin{equation}
\mathcal{L}_{R} = \sum_{f_g \in \mathcal{F}}\sum_{t=2}^{T}\frac{1}{\gamma^t}\mathrm{ReLU}(y_{opt} - \max_{i\leq t} y_i),
\label{eq:l_r}
\end{equation}
where $\mathrm{ReLU}(x) = \max(x,0)$.
Since the regret is expected to be high during initial iterations because of random initialization of $\mathbf{x}$ but desired to be low close to $T$, we give exponentially increasing importance to regret terms via a discount factor $0<\gamma \leq 1$. 
In contrast to regret loss, OI loss used in RNN-OI is given by \cite{chen2017learning,zhou2017optimizing}: 

\begin{equation}
	\mathcal{L}_{OI} = \sum_{f_g \in \mathcal{F}}\sum_{t=2}^{T}\frac{1}{\gamma^t}\mathrm{ReLU}(y_t - \max_{i<t} y_i )
	\label{eq:loss_oi}
\end{equation}

It is to be noted that using $\mathcal{L}_R$ as the loss function mimics a supervised scenario where the target $y_{opt}$ for each optimization task is known and explicitly used to guide the learning process. On the other hand, $\mathcal{L}_{OI}$ mimics an unsupervised scenario where the target $y_{opt}$ is unknown and the learning process solely relies on the feedback about whether it is able to improve $y_t$ over iterations. It is important to note that once trained, the model requires neither $y_{opt}$ nor $\mathbf{x}_{opt}$ during inference.

\subsubsection{Incremental Normalization\label{ssec:in}}
We do not assume any constraint on the range of values the functions $f_g$ and $f_b$ can take. 
Although this feature is critical for most practical aspects, it poses a challenge on the training and inference procedures using RNN:
Neural networks are known to work well only on normalized inputs, and can be numerically unstable and difficult to train on very large or very small values as typical non-linear activation functions like sigmoid activation function tend to saturate for large inputs and will adjust slowly during training. 
RNNs are most easily trained when their inputs are well conditioned, and have a similar scale as their latent state, and suitable scaling often accelerates training \cite{ioffe2015batch,wichrowska2017learned}.
This poses a challenge during both training and inference if we directly use $y_t$ as an input to the RNN. \begin{figure}
	\centering
	\includegraphics[scale=0.2]{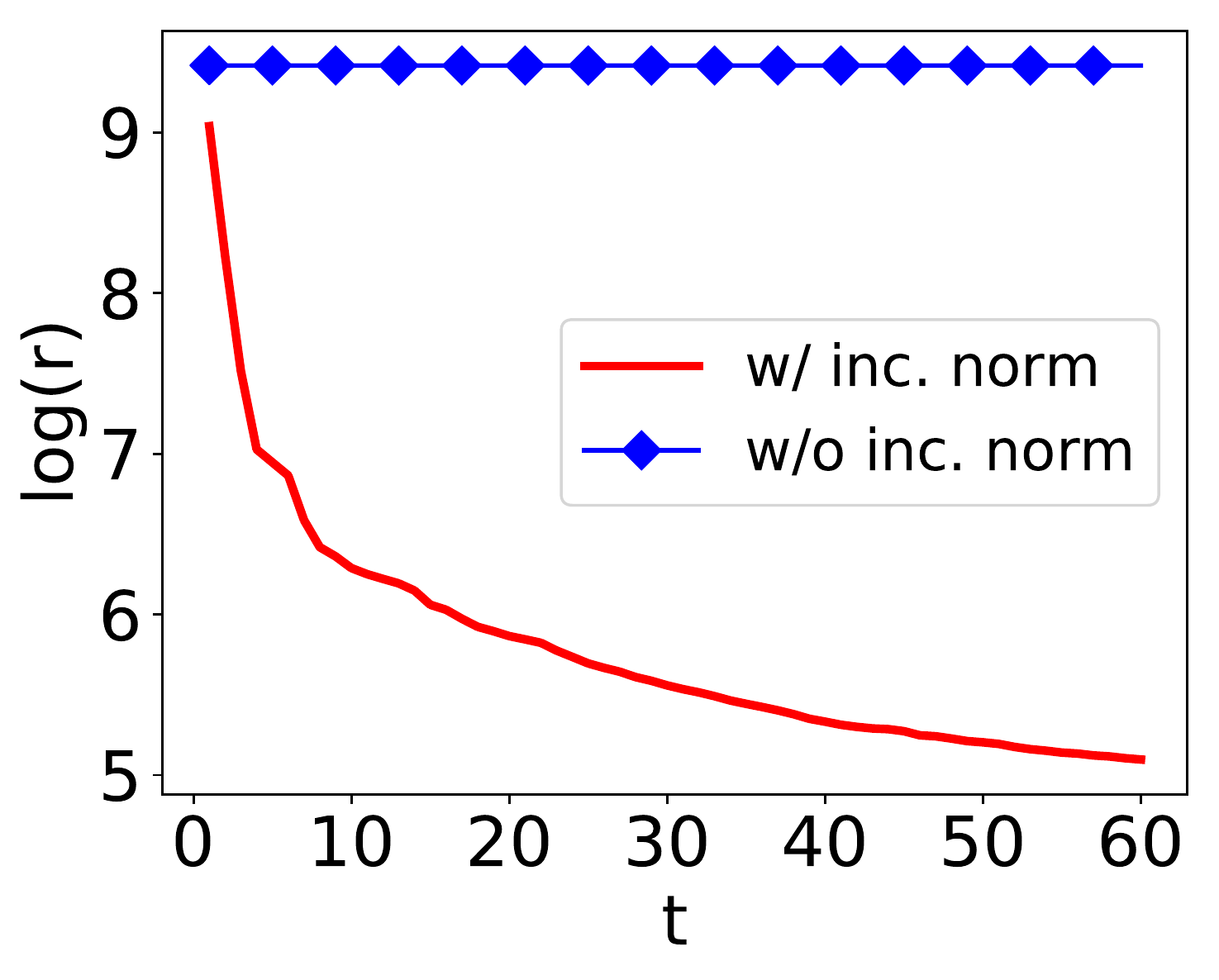}
	\caption{Effect of not using suitable scaling (incremental normalization in our case) of black-box function value during inference.\label{fig:inc-norm-test}}
\end{figure}
Fig. \ref{fig:inc-norm-test} illustrates the saturation effect if suitable incremental normalization of function values is not used during inference. This behavior at inference time was noted\footnote{as per electronic correspondence with the authors} in \cite{chen2017learning}, however, was not considered while training RNN-OI.
In order to deal with any range of values that $f_g$ can take during training or that $f_b$ can take during inference, we consider incremental normalization while training such that $y_{t}$ in Eq. \ref{eq:rnn-opt} is replaced by $\tilde y_{t} = \frac{y_t - \mu_t}{\sqrt{\sigma_t^2 + \epsilon}}$ such that $\mathbf{h}_{t+1} = f_o(\mathbf{h}_{t}, \mathbf{x}_{t}, \tilde y_{t};\boldsymbol{\theta})$,
where $\mu_t = \frac{1}{t}\sum_{i=1}^{t}y_i$, $\sigma_t^2 = \frac{1}{t}\sum_{i=1}^{t}(y_i-\mu_t)^2$, and $0<\epsilon \ll 1$. (We used $\epsilon=0.05$ in our experiments).

\subsection{RNN-Opt with Domain Constraints (RNN-Opt-DC)}
Consider a constrained optimization problem of finding $\mathrm{arg}\max_\mathbf{x} f_b(\mathbf{x})$ subject to constraints given by $c_j(\mathbf{x})\leq 0, ~ j=1,\ldots,C$, where $C$ is the number of constraints.
To ensure that the optimizer proposes queries that satisfy the domain constraints, or is at least able to receive feedback when it proposes a query that violates any domain constraints, we consider the following enhancements in RNN-Opt, as depicted in Fig. \ref{fig:rnn-opt-dc}: 
\begin{figure}
	\centering
	\includegraphics[scale=0.35,trim={0cm, 2.5cm, 0cm, 0cm},clip]{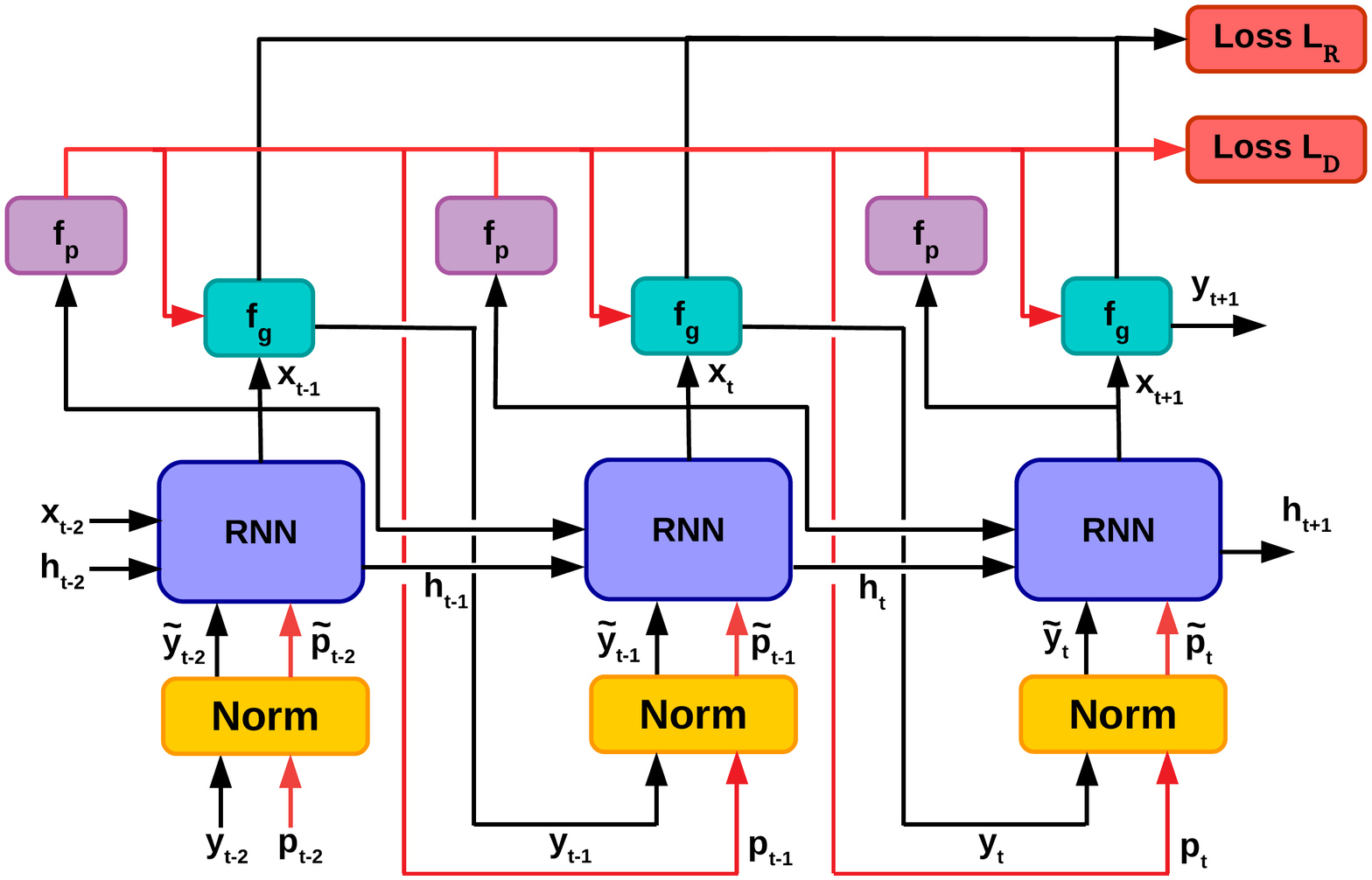}
	\caption{Computation flow in RNN-Opt-DC. Here $f_g$ is the function to be optimized, and $f_p$ is used to compute the penalty $p_t$. Further, if $p_t=0$, actual value of $f_g$, i.e. $y_t$ is passed to the loss function and RNN, else $y_t$ is set to $y_{t-1}$. \label{fig:rnn-opt-dc}}
\end{figure}

1. Input an explicit feedback $p_t$ via a penalty function s.t. $p_t = f_p(\mathbf{x}_t)$ to the RNN that captures the extent to which a proposed query $\mathbf{x}_t$ violates any of the $C$ domain constraints. We consider the following instantiation of penalty function: $f_p(\mathbf{x}_t) = \sum_{j=1}^{C} \mathrm{ReLU}(c_j(\mathbf{x}_t))$, i.e. for any $j$ for which $c_j(\mathbf{x}_t) >0$ a penalty equal to $c_j(\mathbf{x}_t)$ is considered, while for any $j$ with $c_j(\mathbf{x}_t) \leq 0$ the contribution to penalty is 0.
The real-valued penalty captures the cumulative extent of violation as well.
Further, similar to normalizing $y_t$, we also normalize $p_t$ incrementally and use $\tilde p_t$ as an additional input to the RNN, such that:
		
\begin{equation}
\mathbf{h}_{t+1} = f_o(\mathbf{h}_{t}, \mathbf{x}_{t}, \tilde y_{t}, \tilde p_t;\boldsymbol{\theta}).
\end{equation}
Further, whenever $p_t>0$, i.e. when one or more of the domain constraints are violated for the proposed query, we set $y_t=y_{t-1}$ rather than actually getting a response from the black-box. This is useful in practice: for example, when trying to optimize a complex dynamical system, getting a response from the system for such a query is not possible as it can be catastrophic.

2. During training, an additional domain constraint loss $\mathcal{L}_D$ is considered that penalizes the optimizer if it proposes a query that does not satisfy one or more of the domain constraints.
\begin{equation}
\mathcal{L}_D = \frac{1}{C}\sum_{f_g \in \mathcal{F}}\sum_{t=2}^{T}p_t.
\label{eq:l_d_binary}
\end{equation}
The overall loss is then given by: 
\begin{equation}
\mathcal{L} = \mathcal{L}_R + \lambda \mathcal{L}_D,
\label{eq:loss-rnn-opt-dc}
\end{equation}
where $\lambda$ controls how strictly the constraints on the domain of parameters should be enforced; higher $\lambda$ implies stricter adherence to constraints. 
It is worth noting that the above formulation of incorporating domain constraints does not put any restriction on the number of constraints $C$ nor on the nature of constraints in the sense that the constraints can be linear or non-linear in nature.  
Further, complex non-linear constraints based on domain knowledge can also be incorporated in a similar fashion during training, e.g. as used in \cite{jia2018physics,muralidhar2018incorporating}. 
Apart from optimizing (in our case, maximizing) $f_g$, the optimizer is also being simultaneously trained to minimize $f_p$.

\subsubsection{Example of penalty function.\label{sssec:dc-example}} Consider simple limit constraints on the input parameters such that the domain of the function $f_g$ is given by $ \Theta = [\mathbf{x}_{min}, \mathbf{x}_{max}]$, then we have:
\begin{equation}
f_p(\mathbf{x}_t) = \sum_{j=1}^{d}\Big (\mathrm{ReLU}(x_t^j - x^j_{max} ) + \mathrm{ReLU}(x^j_{min} - x_t^j )\Big ),
\label{eq:l_d_relu}
\end{equation}
where $x_t^j$ denotes the $j$-th dimension of $\mathbf{x}_t$ where $ x^j_{min}$ and $ x^j_{max}$ are the $j$-th elements of $\mathbf{x}_{min}$ and $\mathbf{x}_{max}$, respectively.

\section{Experimental Evaluation\label{sec:exp}}
We conduct experiments to evaluate the following: i. regret loss ($\mathcal{L}_R$) versus OI loss ($\mathcal{L}_{OI}$), ii. effect of including incremental normalization during training, and iii. ability of RNN-Opt trained with domain constraints using $\mathcal{L}$ (Eq. \ref{eq:loss-rnn-opt-dc}) to generate more feasible queries and leverage feedback to quickly adapt in case it proposes queries violating domain constraints.

For the unconstrained setting, we test RNN-Opt on i) standard benchmark functions for $d=2$ and $d=6$, and ii) 1280 synthetically generated GMM-DF functions (refer Appendix \ref{apx:gmm-df}) not seen during training. 
We choose the benchmark functions such as Goldstein, Rosenbrock, and Rastrigin (and the simple spherical function) that are known to be challenging for standard optimization methods. None of these functions were used for training any of the optimizers.

We use regret $r_t = y_{opt} - \max_{i\leq t} y_i$ to measure the performance of any optimizer after $t$ iterations, i.e. after proposing $t$ queries. Lower values of $r_t$ indicate superior optimizer performance. We test all the optimizers under limited budget setting such that $T=10\times d$.
For each test function, the first query is randomly sampled from $U(-4.0,4.0)$, and we report average regret $r_t$ over 1280 random initializations. 
For synthetically generated GMM-DF functions, we report average regret over 1280 functions with one random initialization for each.

All RNN-based optimizers (refer Table \ref{tab:opt-variants}) were trained for 8000 iterations using Adam optimizer \cite{kingma2014adam} with an initial learning rate of 0.005. The network consists of two hidden layers with the number of LSTM units in each layer being chosen from $\{80,120,160\}$ using a hold-out validation set of $1280$ GMM-DF. Another set of 1280 randomly generated functions constitute the GMM-DF test set. An initial code base\footnote{\url{https://github.com/lightingghost/chemopt}} developed using Tensorflow \cite{abadi2016tensorflow} was adapted to implement our algorithm. 
We used a batch size of 128, i.e. 128 randomly-sampled functions (refer Equation \ref{eq:gmm}) are processed in one mini-batch for updating the parameters of LSTM.

\begin{table}
	\centering
	\scriptsize
	\caption{Variants of trained optimizers considered\label{tab:opt-variants}. Each row corresponds to a method. Y/N denote whether a feature (incremental normalization or domain constraint) was considered (Y) or not (N) during training or inference in a particular method.}
	\begin{tabular}{|c|c|c|c|c|c|c|}
		\hline
		Method & Loss & $\gamma$ & \multicolumn{2}{|c|}{Inc. Norm.} & \multicolumn{2}{|c|}{Domain Const. (DC)}\\
		\hline
		&&&Training&Inference&Training&Inference\\
		\hline
		RNN-OI&$\mathcal{L}_{OI}$&1.0&N&Y&N&N\\
		\hline
		RNN-Opt-Basic&$\mathcal{L}_{R}$&0.98&N&Y&N&N\\
		\hline
		RNN-Opt&$\mathcal{L}_{R}$&0.98&Y&Y&N&N\\
		\hline
		RNN-Opt-P&$\mathcal{L}_{R}$&0.98&Y&Y&N&Y\\
		\hline
		RNN-Opt-DC&$\mathcal{L}_{R}+\lambda \mathcal{L}_{D}$&0.98&Y&Y&Y&Y\\
		\hline
	\end{tabular}
\end{table}

\subsection{Observations}
\begin{figure}
	\centering
	\subfigure[\scriptsize GMM-DF (d=2)]{\includegraphics[width=0.24\textwidth]{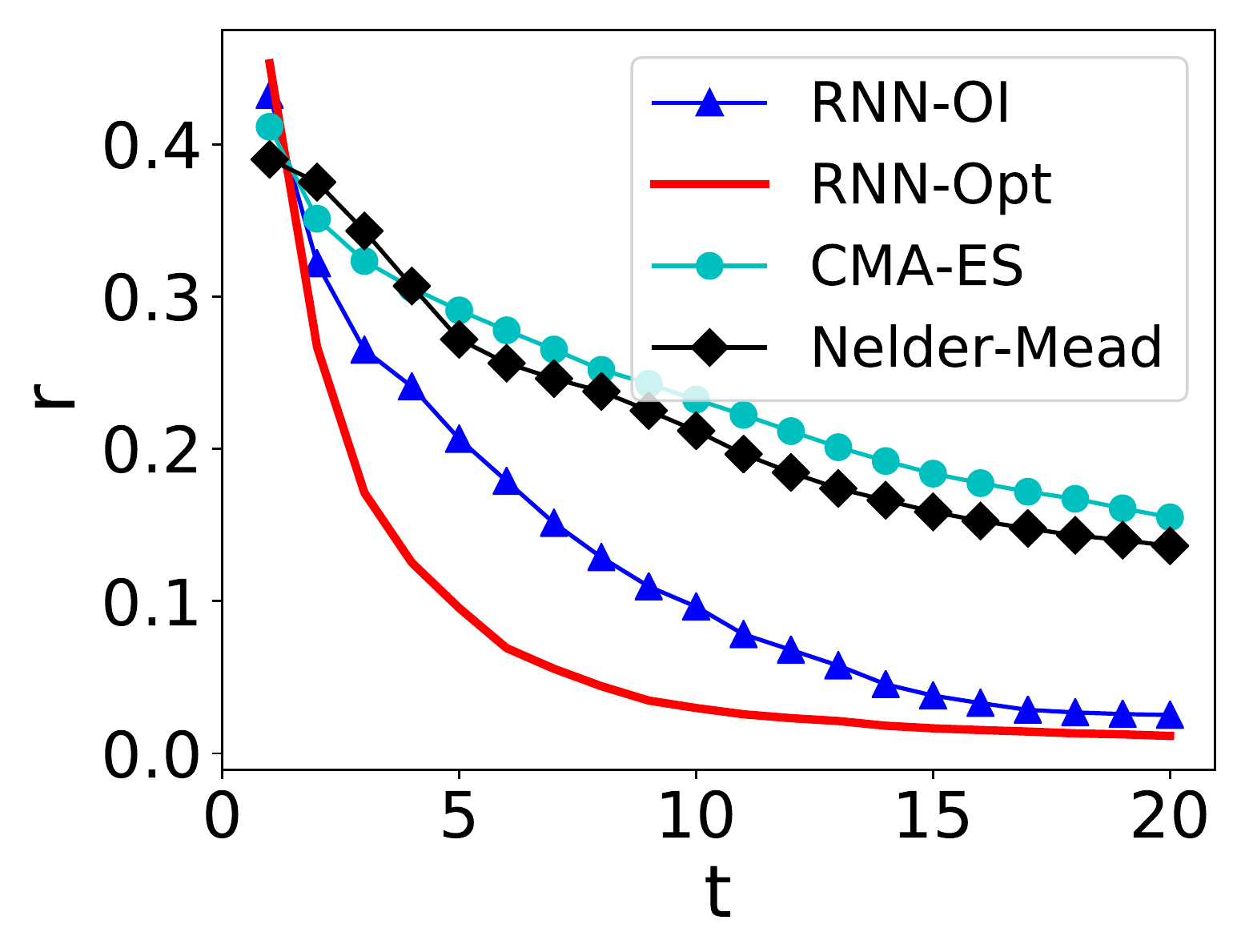}}
	\subfigure[\scriptsize Goldstein (d=2)]{\includegraphics[width=0.24\textwidth]{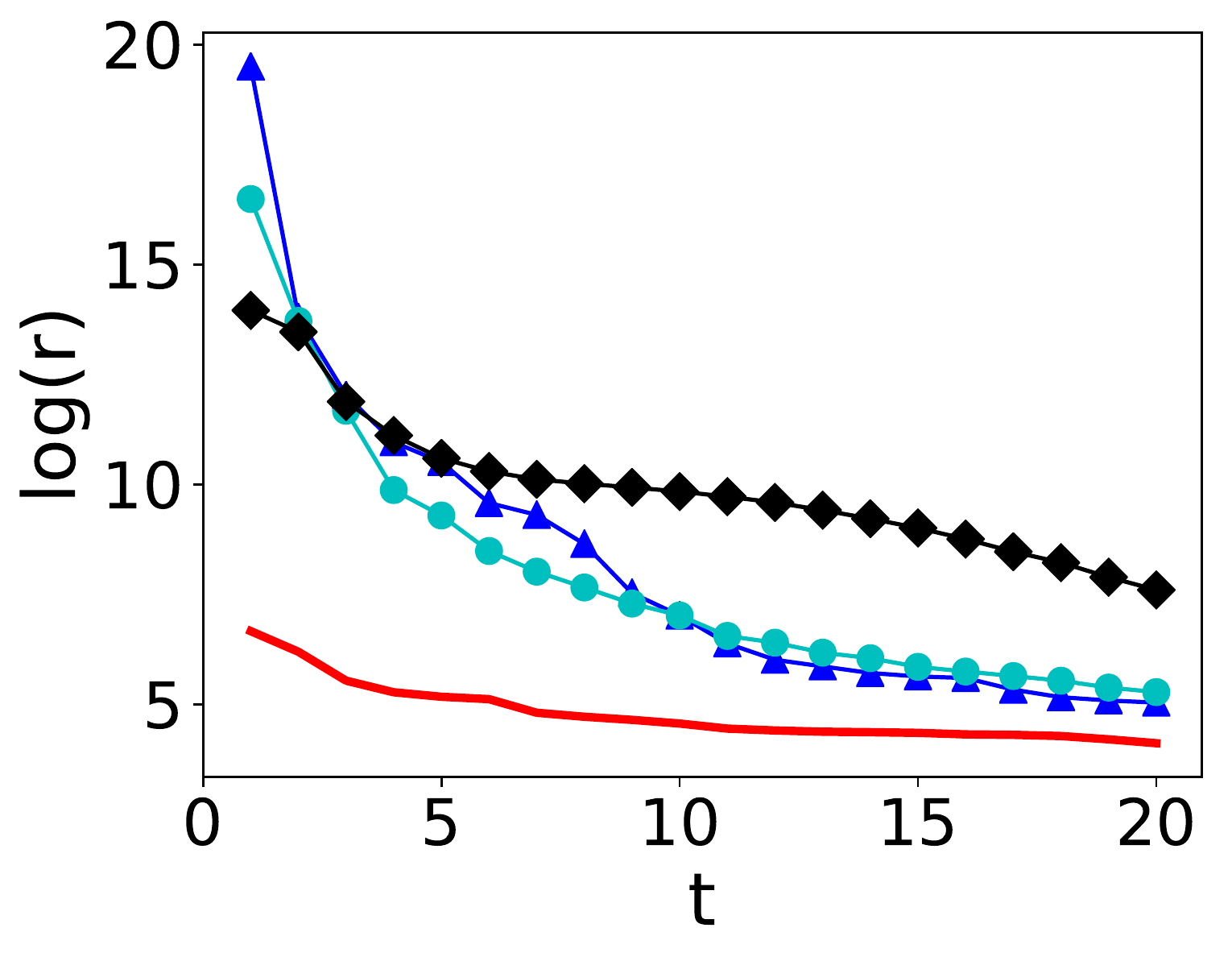}}
	\subfigure[\scriptsize Rastrigin (d=2)]{\includegraphics[width=0.24\textwidth]{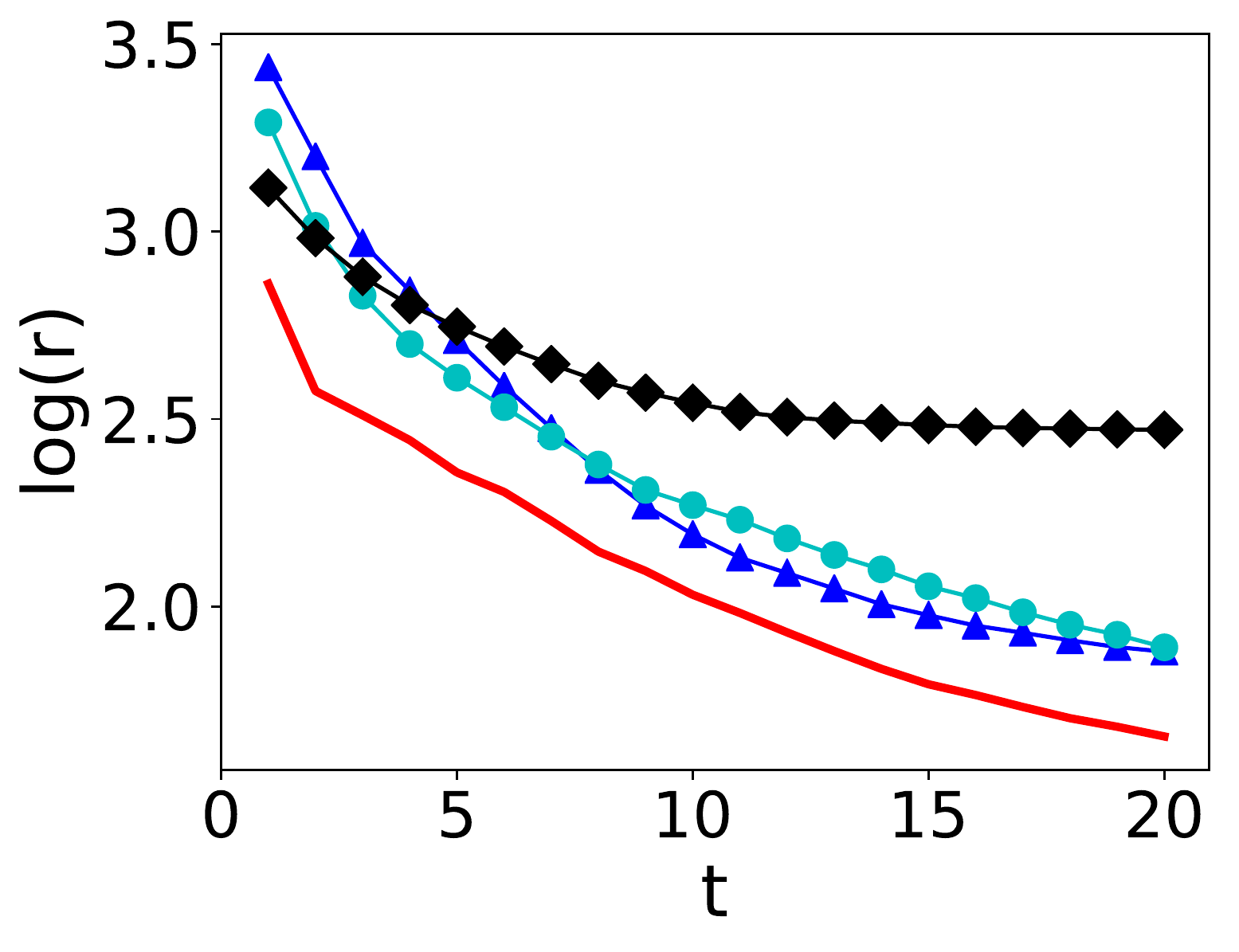}}
	\subfigure[\scriptsize Rosenbrock (d=2)]{\includegraphics[width=0.24\textwidth]{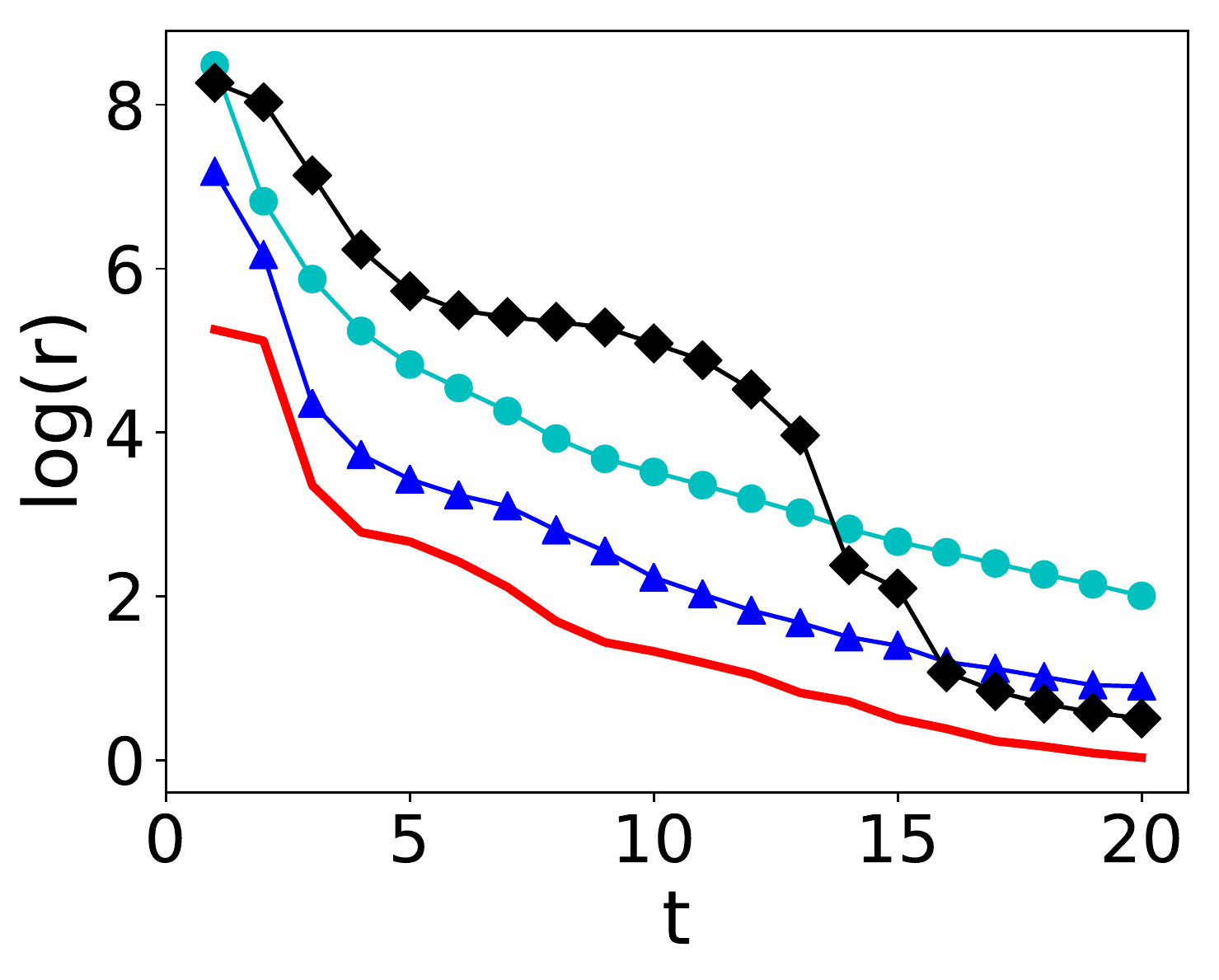}}
	\subfigure[\scriptsize Spherical (d=2)]{\includegraphics[width=0.24\textwidth]{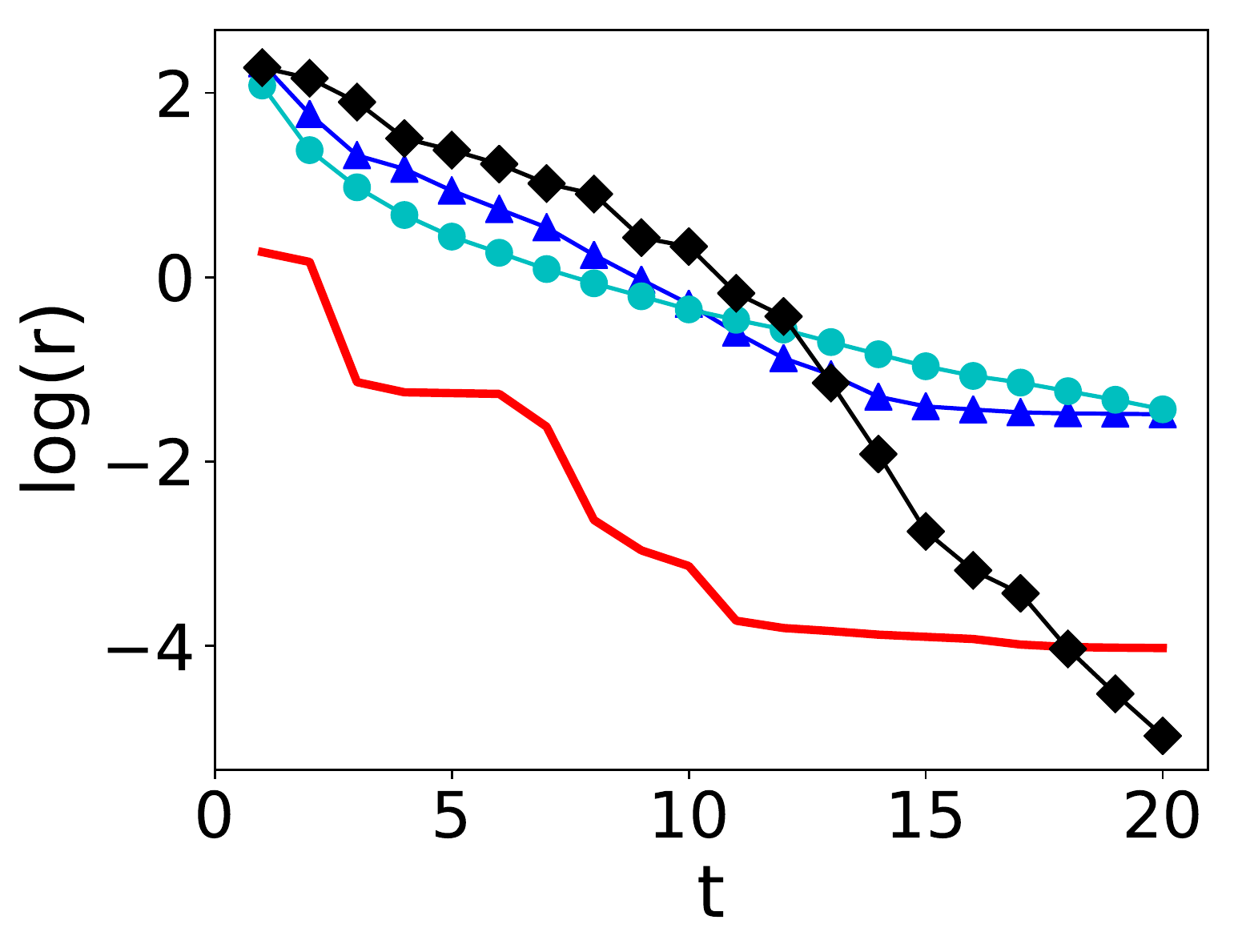}}
	\subfigure[\scriptsize GMM-DF (d=6)]{\includegraphics[width=0.24\textwidth]{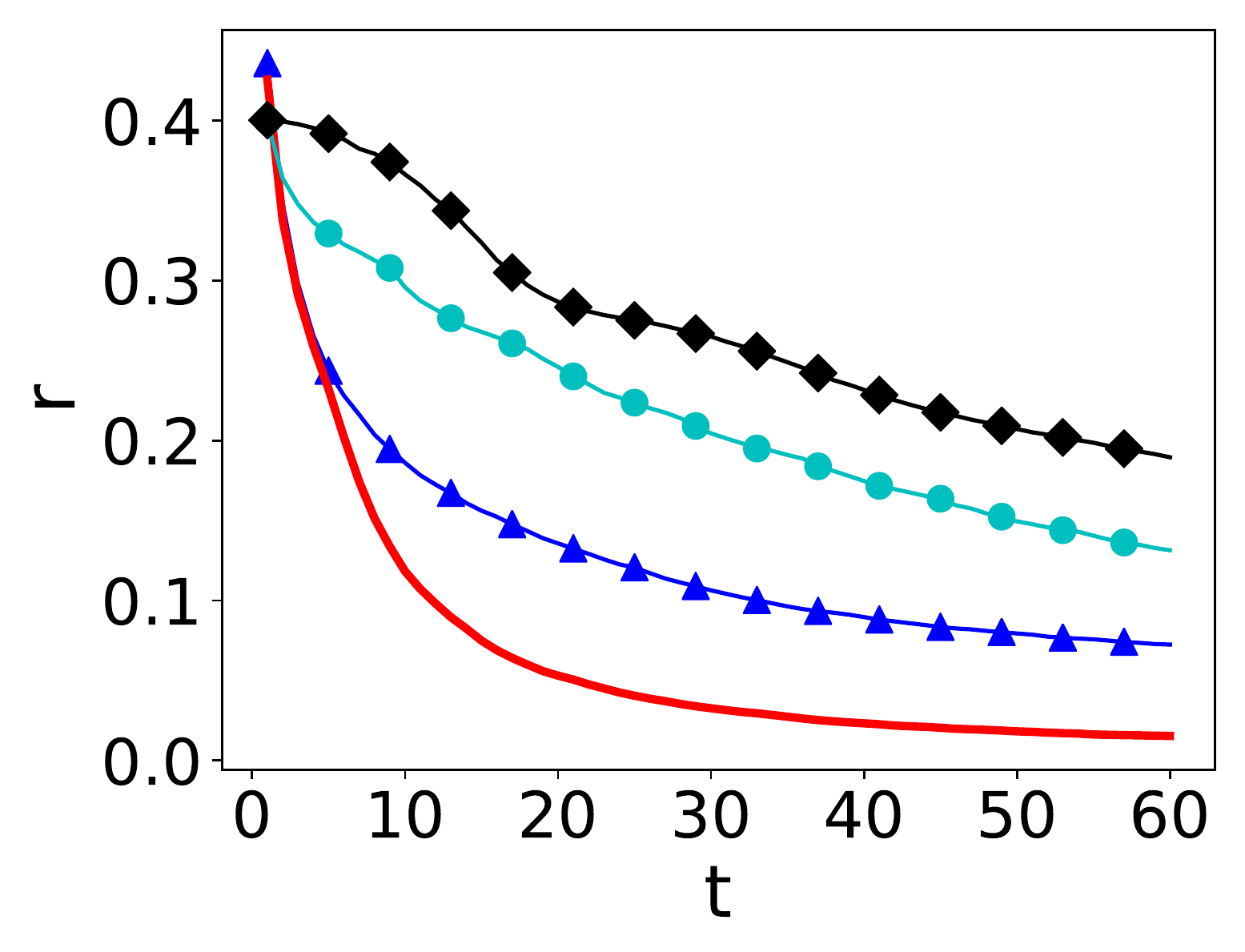}}
	\subfigure[\scriptsize Rastrigin (d=6)]{\includegraphics[width=0.24\textwidth]{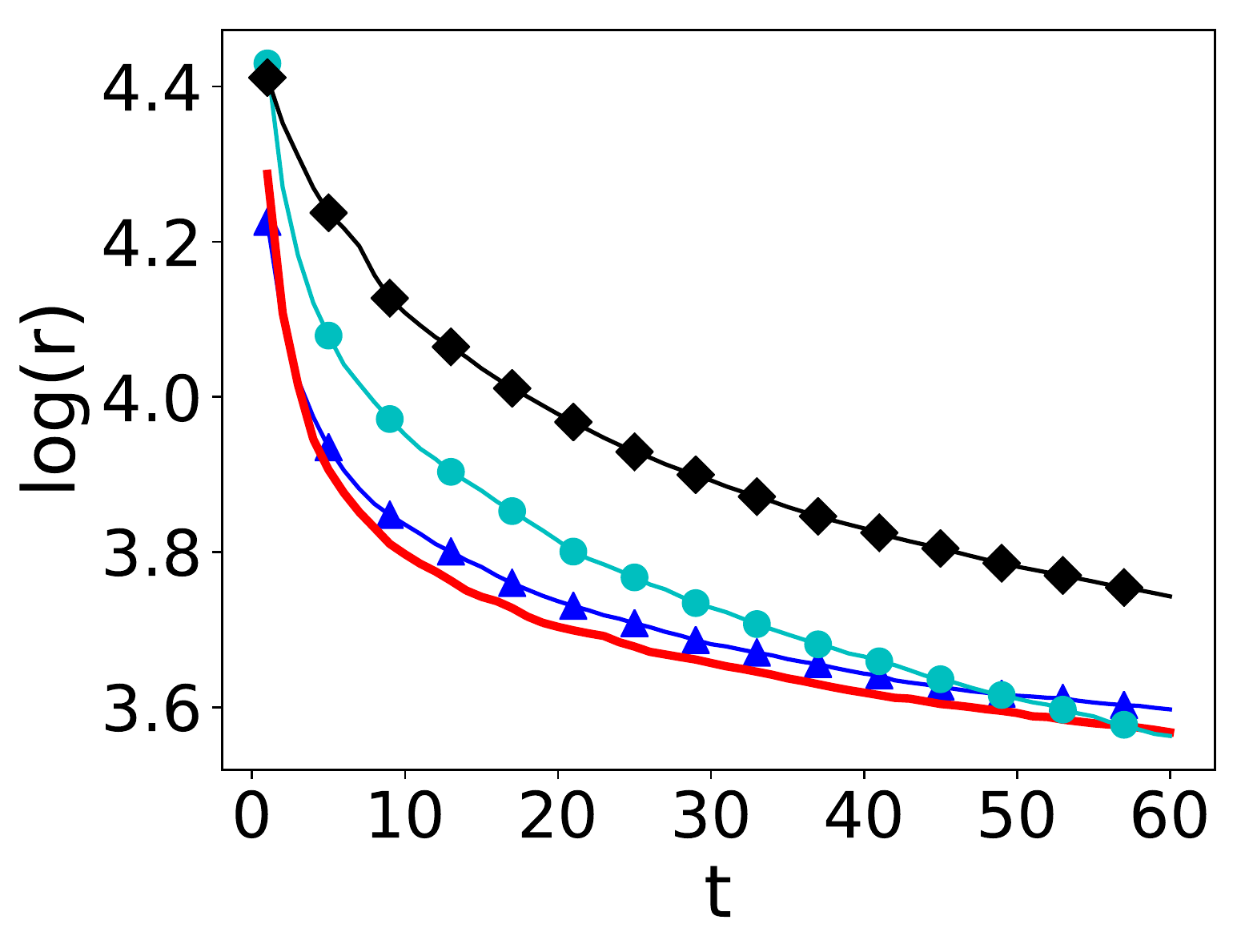}}
	\subfigure[\scriptsize Rosenbrock (d=6)]{\includegraphics[width=0.24\textwidth]{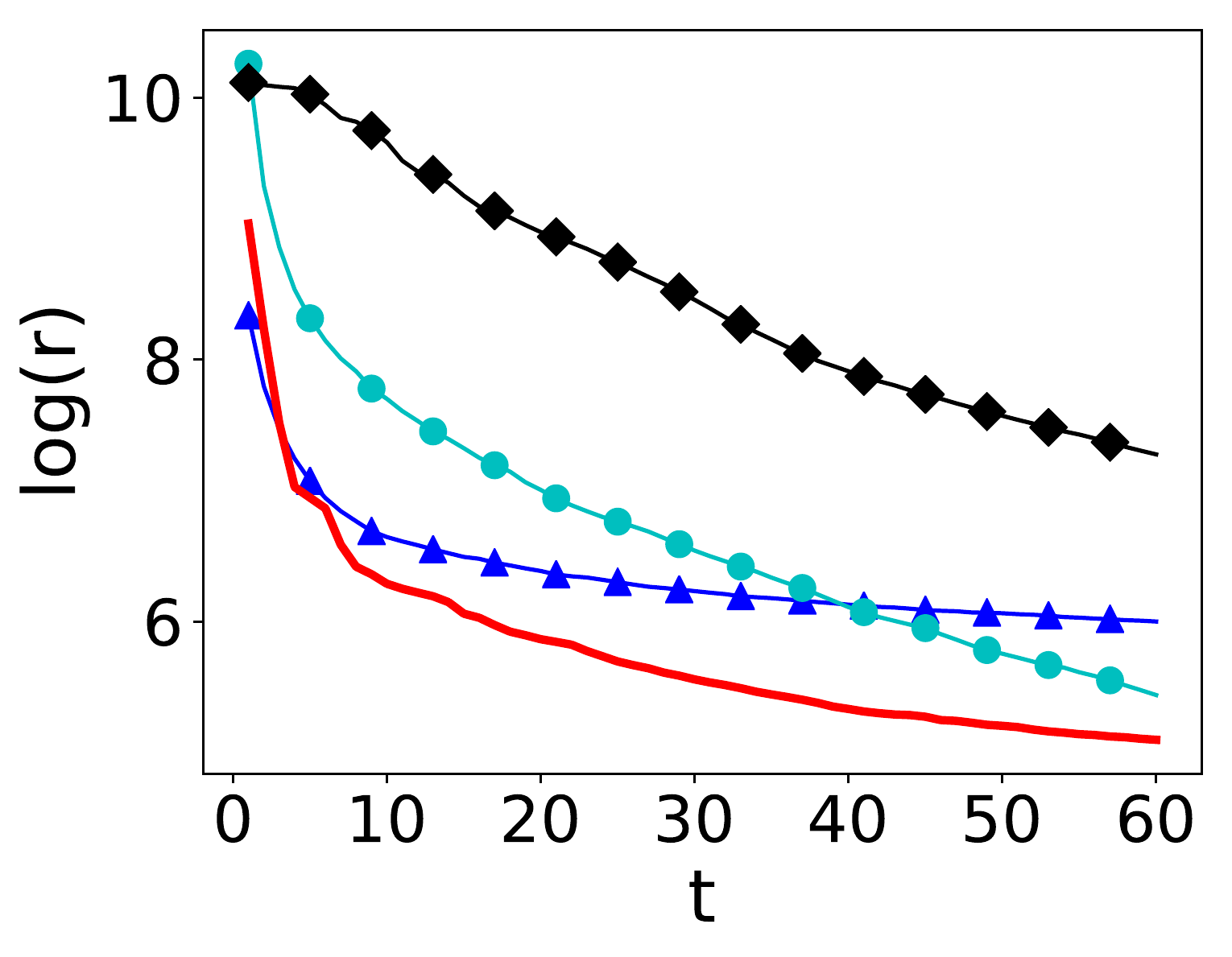}}
	\subfigure[\scriptsize Spherical (d=6)]{\includegraphics[width=0.24\textwidth]{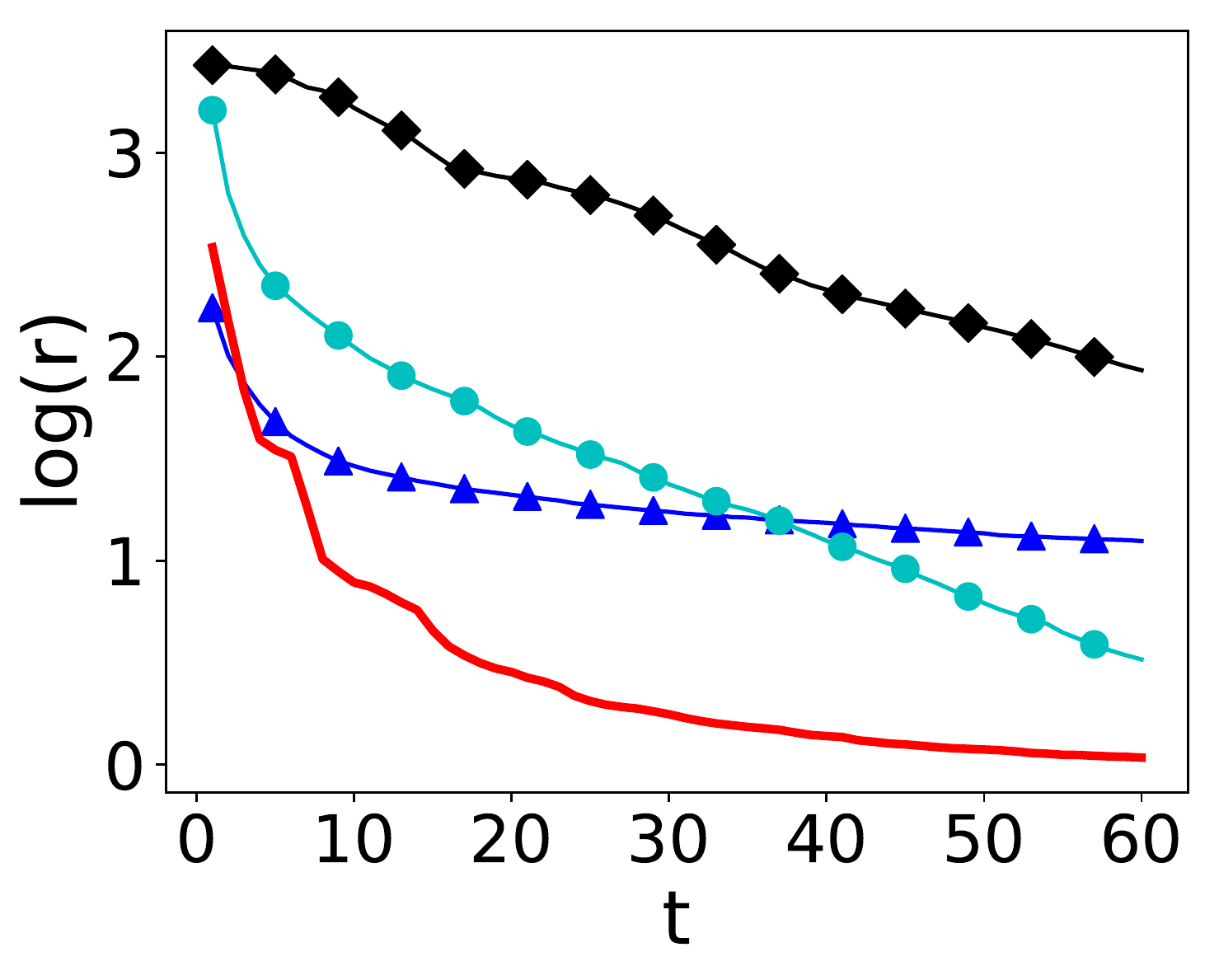}}
	\subfigure[\scriptsize GMM-DF (d=2)]{\includegraphics[width=0.25\textwidth]{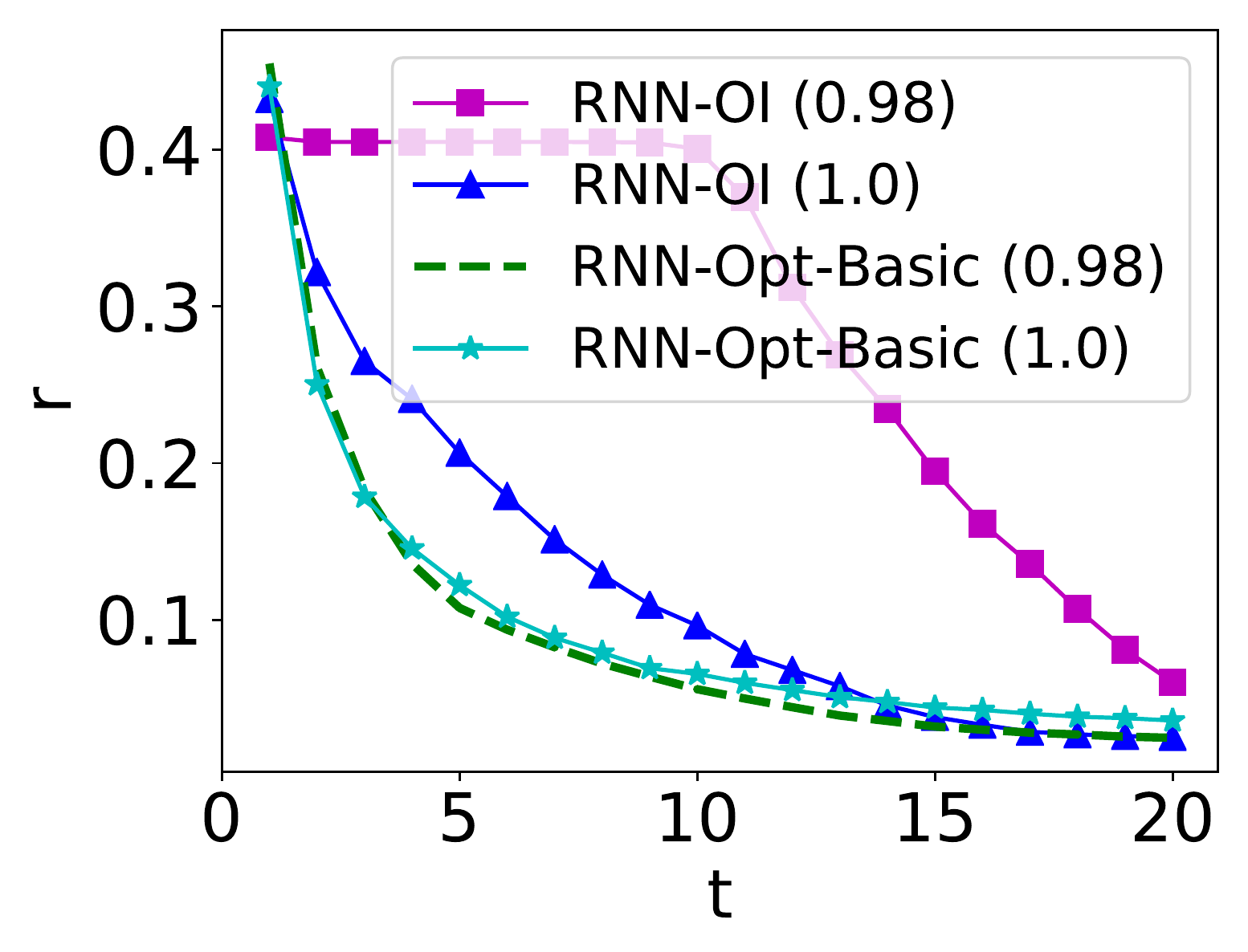}}
	\subfigure[\scriptsize GMM-DF (d=6)]{\includegraphics[width=0.25\textwidth]{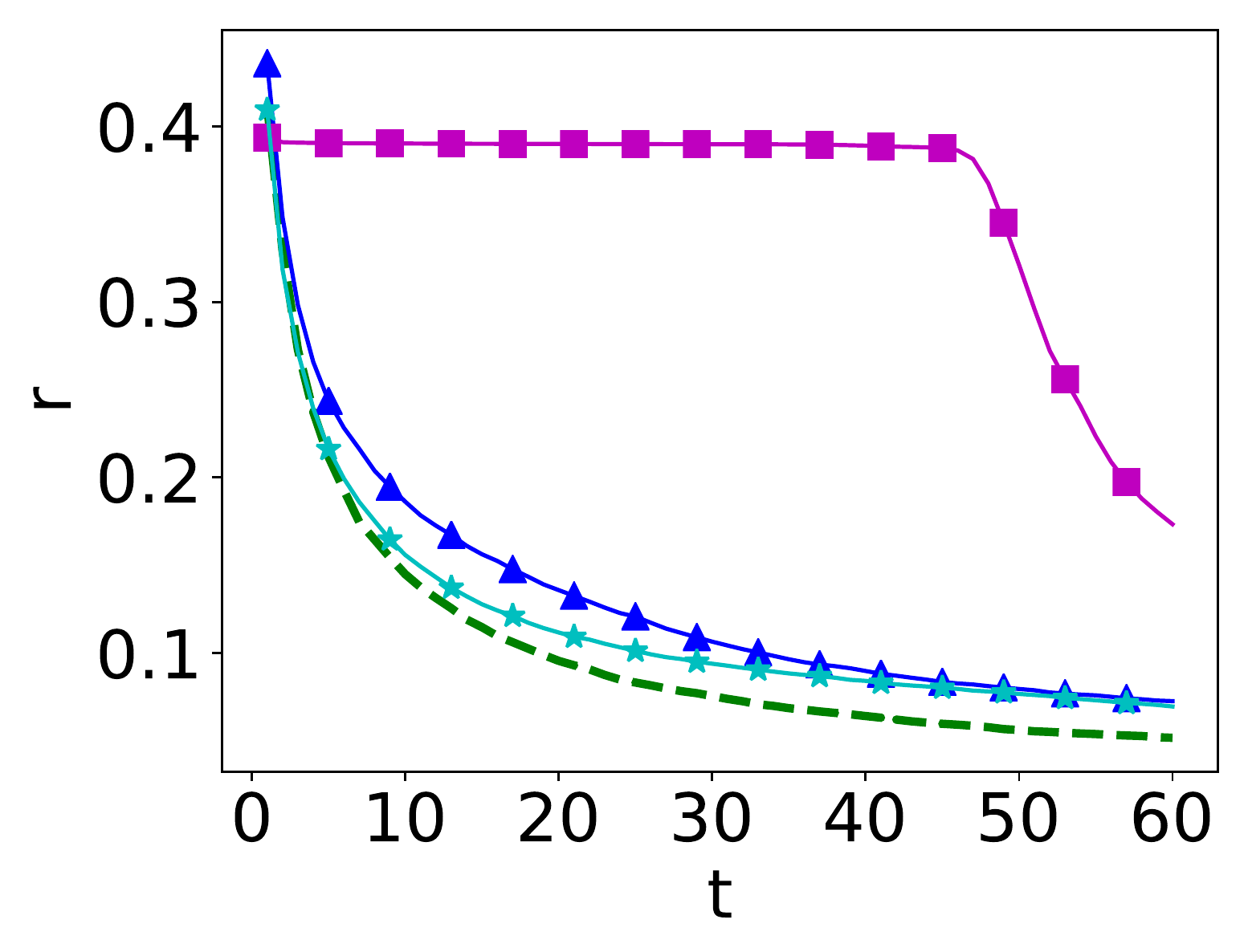}}
		
	\caption{(a)-(i) RNN-Opt versus CMA-ES, Nelder-Mead and RNN-OI for benchmark functions for $d=2$ and $d=6$. (j)-(k) Regret loss versus OI Loss with varying discount factor $\gamma$ mentioned in brackets in the legend. (Lower regret is better.) \label{fig:main}}
\end{figure}

We make the following key observations for unconstrained optimization setting:

1. \textbf{RNN-Opt is able to optimize black-box functions not seen during training, and hence, generalize.} 
We compare RNN-Opt with RNN-OI and two standard black-box optimization algorithms CMA-ES \cite{hansen1996adapting} and Nelder-Mead \cite{nelder1965simplex}. 
RNN-OI uses $\mathbf{x}_t$, $y_t$, and $\mathbf{h}_t$ to get the next hidden state $\mathbf{h}_{t+1}$, which is then used to get $\mathbf{x}_{t+1}$ (as in Eq \ref{eq:gen_x}), such that $\mathbf{h}_{t+1} = f_o(\mathbf{h}_{t}, \mathbf{x}_{t}, y_{t};\boldsymbol{\theta}),\label{eq:RNN-OI}$ with OI loss as given in Eq. \ref{eq:loss_oi}.
From Fig. \ref{fig:main} (a)-(i), we observe that RNN-Opt outperforms all the baselines considered on most functions considered while being at least as good as the baselines in few remaining cases. Except for the simple convex spherical function, RNN-based optimizers outperform CMA-ES and Nelder-Mead under limited budget, i.e. with $T=20$ for $d=2$ and $T=60$ for $d=6$. We observe that trained optimizers outperform CMA-ES and Nelder-Mead for higher-dimensional cases ($d=6$ here, as also observed in \cite{chen2017learning,zhou2017optimizing}).

2. \textbf{Regret-based loss is better than the OI loss.}
We compare \textit{RNN-Opt-Basic} with RNN-OI (refer Table \ref{tab:opt-variants}) where RNN-Opt-Basic differs from RNN-OI only in the loss function (and the discount factor, as discussed in next point).
For fair comparison with RNN-OI, RNN-Opt-Basic does not include incremental normalization during training. 
From Fig. \ref{fig:main} (j)-(k), we observe that RNN-Opt-Basic (with $\gamma=0.98$) performs better than RNN-OI during initial steps for $d=2$ (while being comparable eventually) and across all steps for $d=6$, proving the advantage of using regret loss over OI loss.

3. \textbf{Significance of discount factor when using regret-based loss versus OI loss}. From Fig. \ref{fig:main} (j)-(k), we also observe that the results of RNN-Opt and RNN-OI are sensitive to the discount factor $\gamma$ (refer Eqs. \ref{eq:l_r} and \ref{eq:loss_oi}). $\gamma<1$ works better for RNN-Opt while $\gamma=1$ (i.e. no discount) works better for RNN-OI. This can be explained as follows: the queries proposed initially (small $t$) are expected to be far from $y_{opt}$ due to random initialization, and therefore, have high initial regret. Hence, components of the loss term for smaller $t$ should be given lower weightage in the regret-based loss. 
On the other hand, during later steps (close to $T$), we would like the regret to be as low as possible, and hence a higher importance should be given to the corresponding terms in the regret-based loss. 
In contrast, RNN-OI is trained to keep improving irrespective of $y_{opt}$, and hence giving equal importance to the contribution of each step to the OI loss works best.

\begin{figure}
	\centering
	\subfigure[\scriptsize GMM-DF (d=2)]{\includegraphics[width=0.24\textwidth]{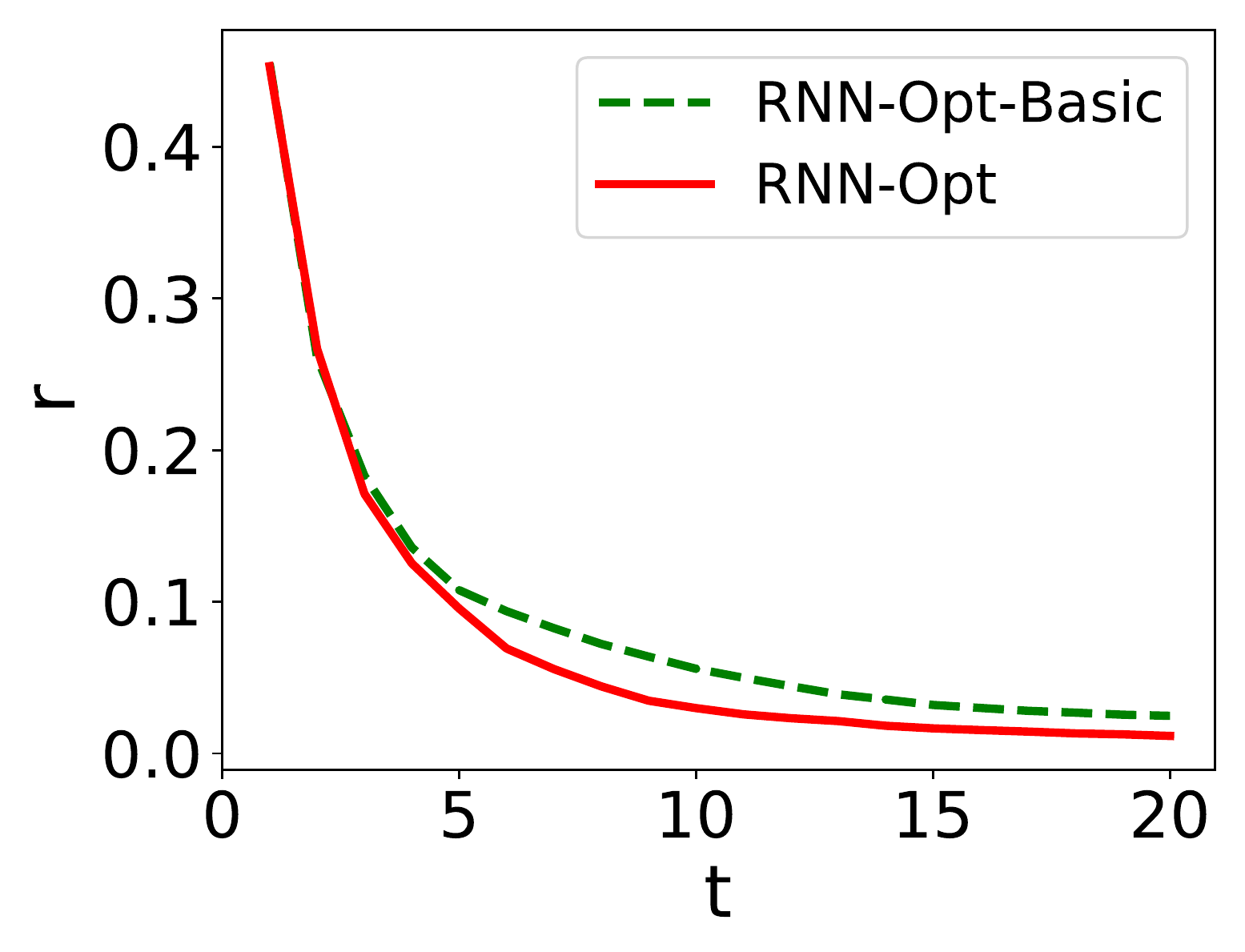}}
	\subfigure[\scriptsize Rosenbrock (d=2)]{\includegraphics[width=0.23\textwidth]{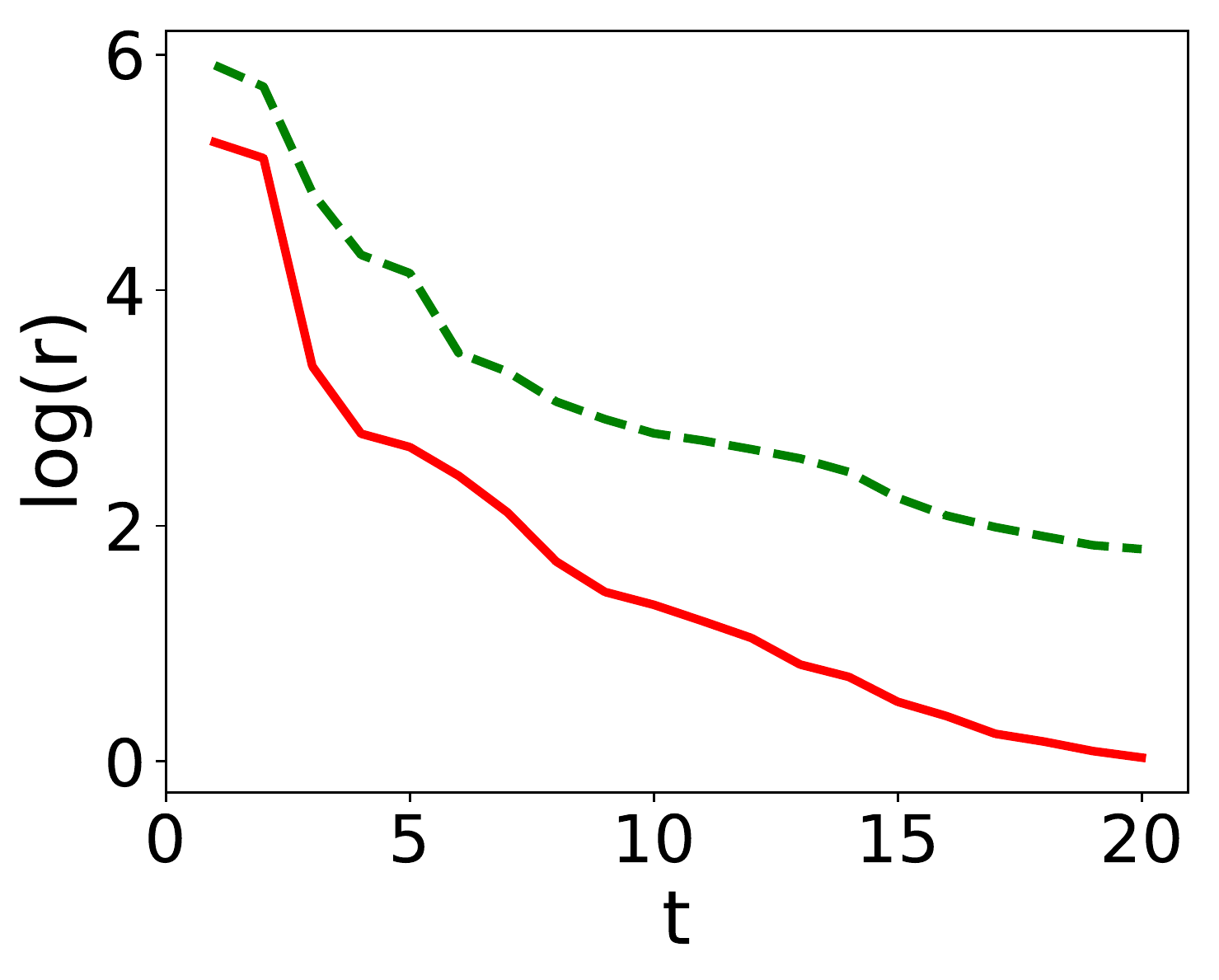}}
	\subfigure[\scriptsize GMM-DF (d=6)]{\includegraphics[width=0.24\textwidth]{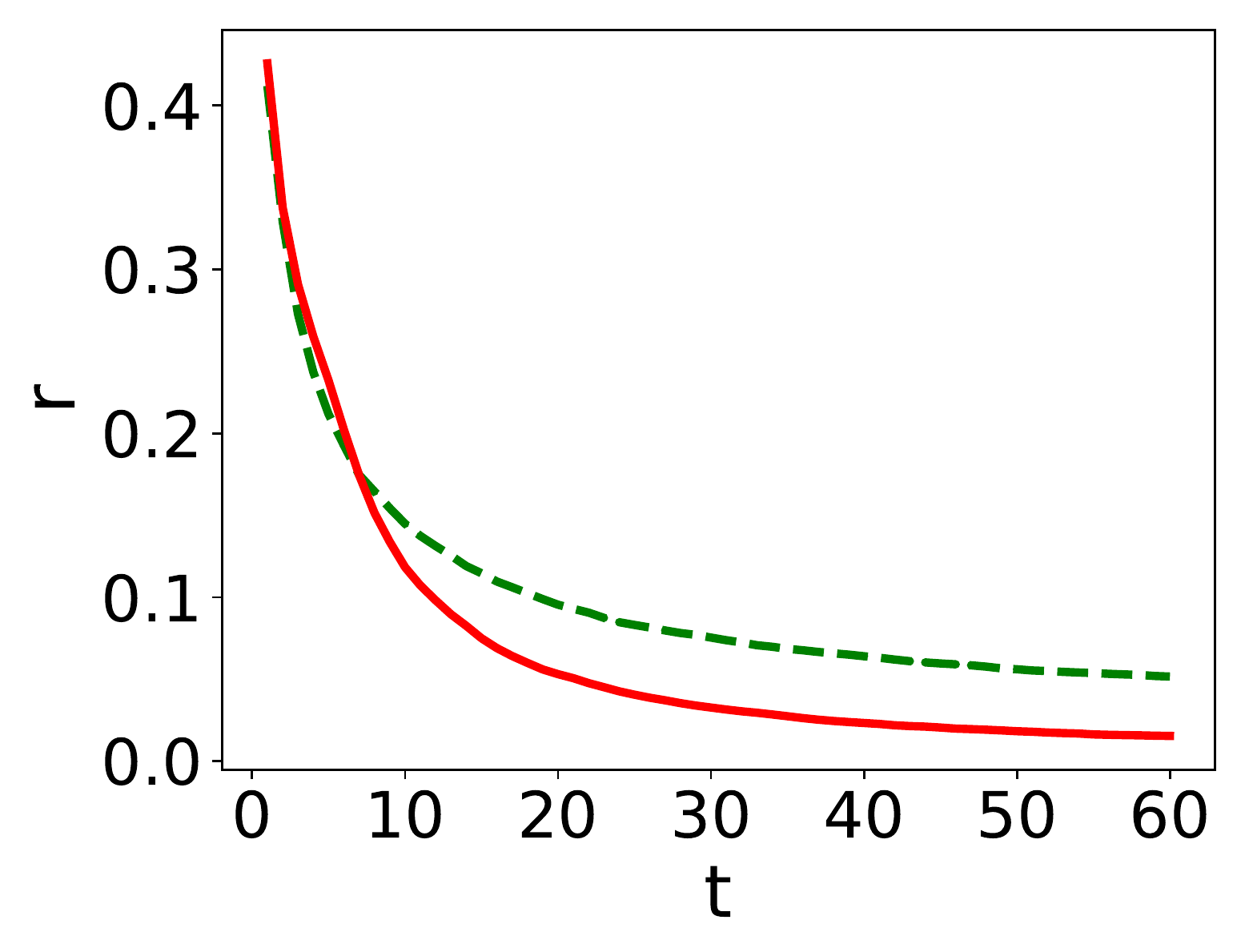}}	
	\subfigure[\scriptsize Rosenbrock (d=6)]{\includegraphics[width=0.23\textwidth]{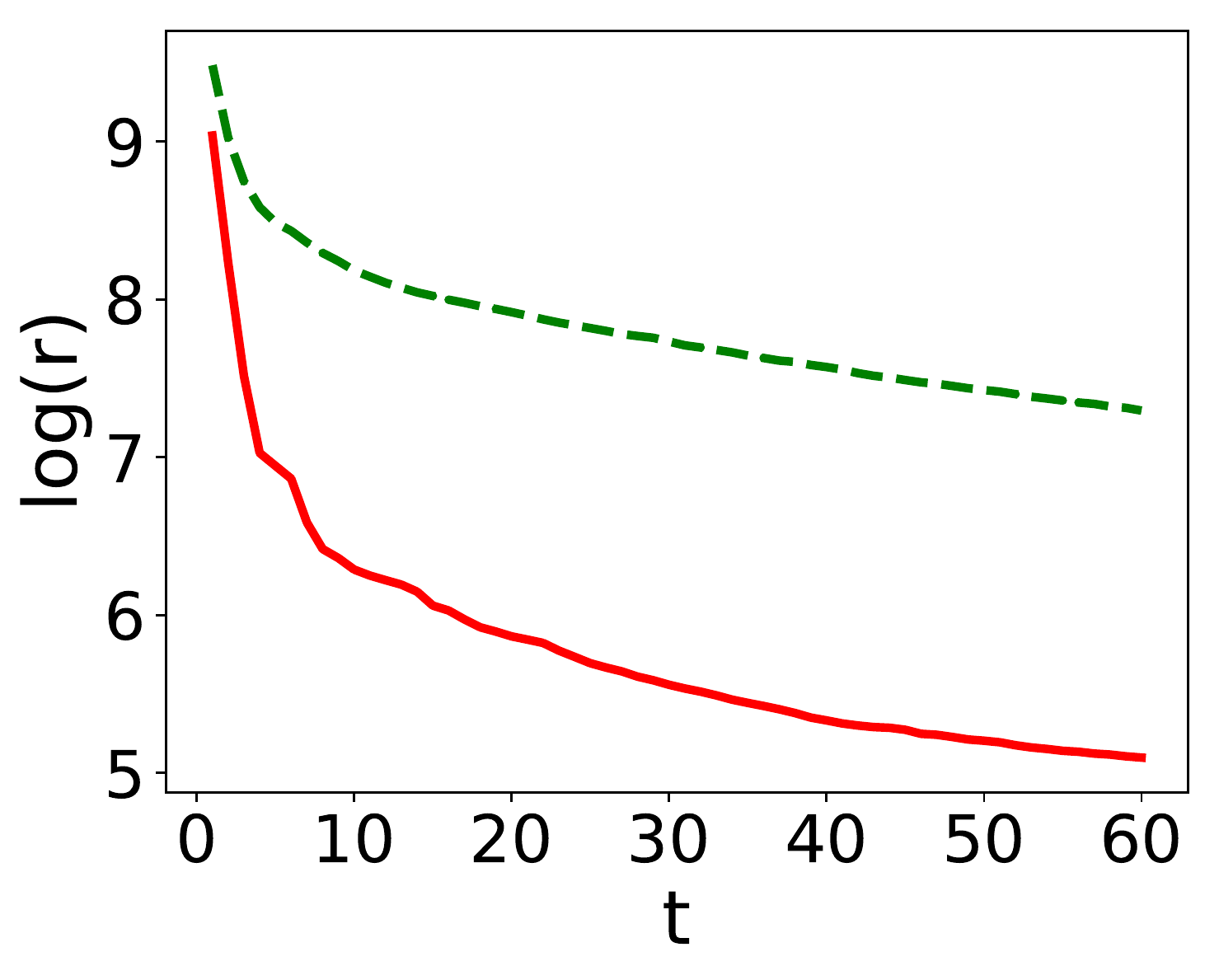}}
	
	\caption{Regret plots showing effect of incremental normalization in RNN-Opt. Similar results are observed for all functions. We omit them here for brevity. \label{fig:inc-norm}}
\end{figure}

4. \textbf{Incremental normalization during training and inference to optimize functions with diverse range of values.} We compare RNN-Opt-Basic and RNN-Opt, where RNN-Opt uses incremental normalization of inputs during training as well as testing (as described in Section \ref{ssec:in}) while RNN-Opt-Basic uses incremental normalization only during testing (refer Table \ref{tab:opt-variants}).
From Fig. \ref{fig:inc-norm}, we observe that RNN-Opt performs significantly better than RNN-Opt-Basic proving the advantage of incorporating incremental normalization during training. Note that since most of the functions considered have large range of values, incremental normalization is by-default enabled for all RNN-based optimizers during testing to obtain meaningful results, as illustrated earlier in Fig. \ref{fig:inc-norm-test}, especially for functions with large range, e.g. Rosenbrock.

\subsection{RNN-Opt with Domain Constraints}
To train RNN-Opt-DC, we generate synthetic functions with random limit constraints as explained in Section \ref{sssec:dc-example}.
The limits of the search space are set as $[\mathbf{x}_{opt}-\Delta \mathbf{x},\mathbf{x}_{opt}+\Delta \mathbf{x}]$ where $\Delta x^j$ ($j$-th component of $\Delta \mathbf{x}$) is sampled from $U(\tau_1,\tau_2)$ (we use $\tau_1=1.0$, $\tau_2=2.0$ during training).

We use $\lambda = 0.2$ for RNN-Opt-DC.
As a baseline, we use RNN-Opt with minor variation during inference time (with no change in training procedure) where, instead of passing $\tilde y_t$ as input to the RNN, we pass $\tilde y_t - \tilde p_t$ so as to capture penalty feedback. We call this baseline approach as \textit{RNN-Opt-P} (refer Table \ref{tab:opt-variants}). While RNN-Opt-DC is explicitly trained to minimize penalty $p_t$ explicitly, RNN-Opt-P captures the requirement of trying to maximize $y_t$ under a soft-constraint of minimizing $p_t$ only during inference time.

\begin{figure}
	\centering
	\subfigure[\scriptsize GMM-DF (d=2)]{\includegraphics[width=0.275\textwidth]{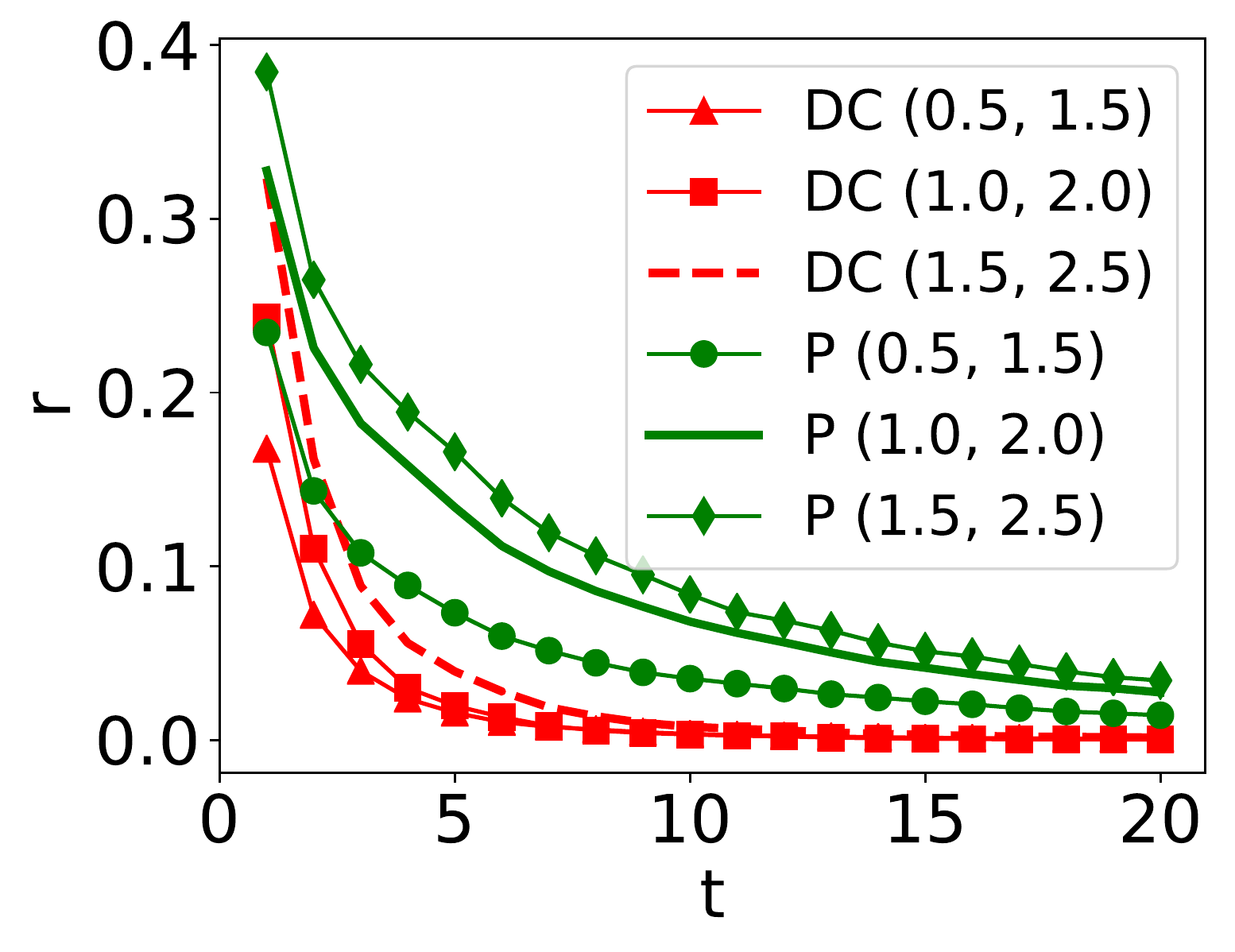}}
	\subfigure[\scriptsize Rosenbrock (d=2)]{\includegraphics[width=0.26\textwidth]{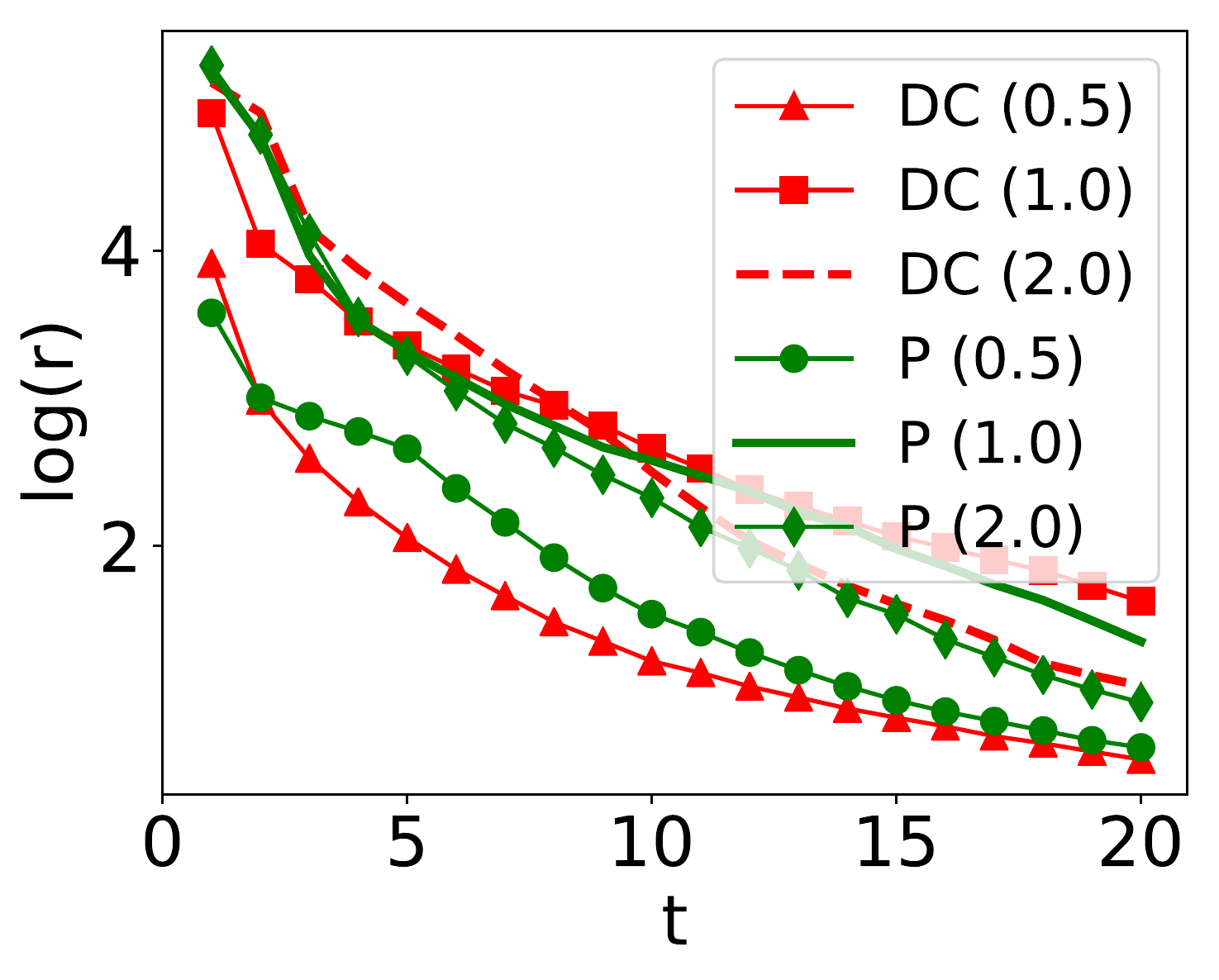}}
	\subfigure[\scriptsize GMM-DF (d=6)]{\includegraphics[width=0.275\textwidth]{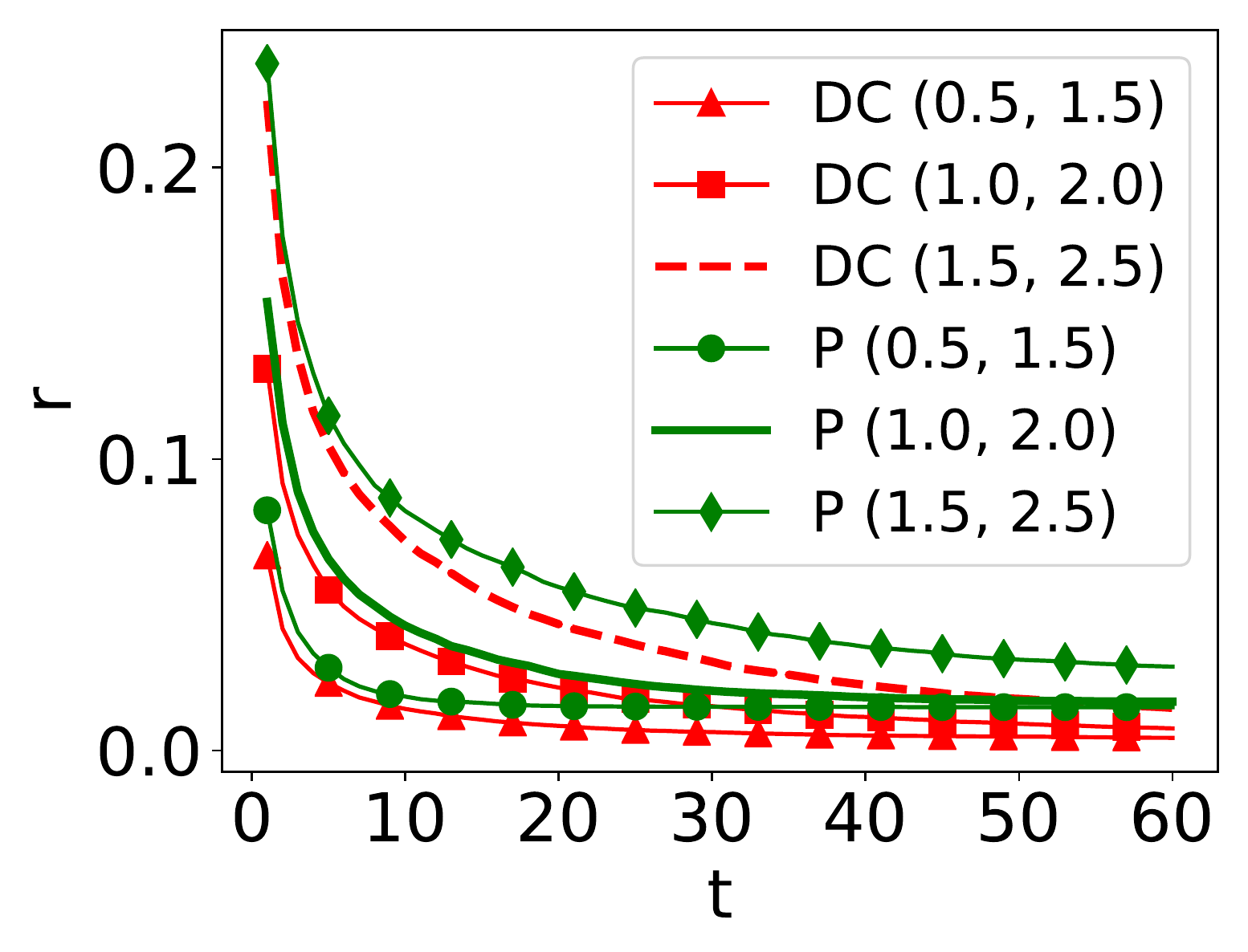}}	
	\subfigure[\scriptsize Rosenbrock (d=6)]{\includegraphics[width=0.26\textwidth]{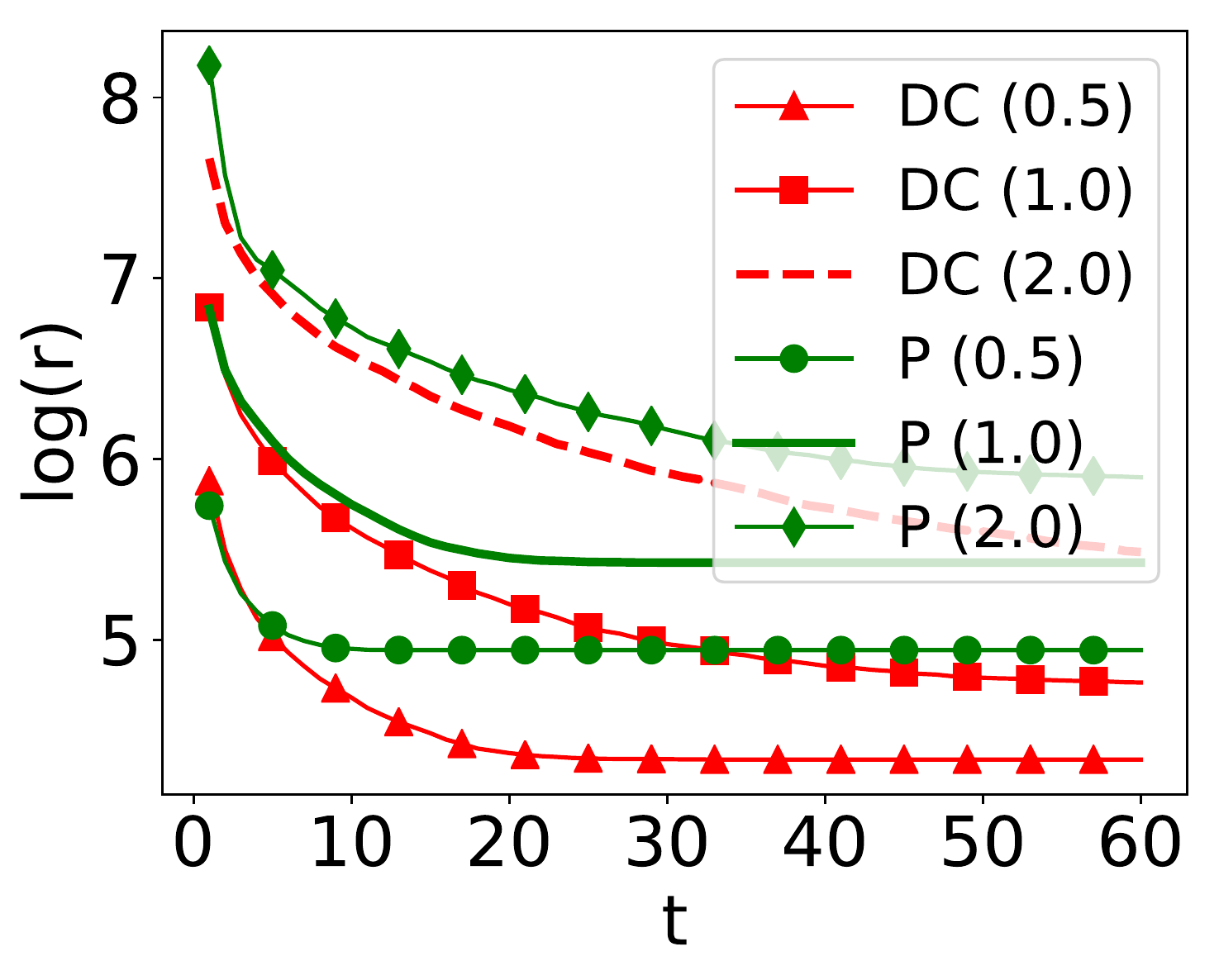}}	
	\subfigure[\scriptsize Industrial-1 (d=6)]{\includegraphics[width=0.26\textwidth]{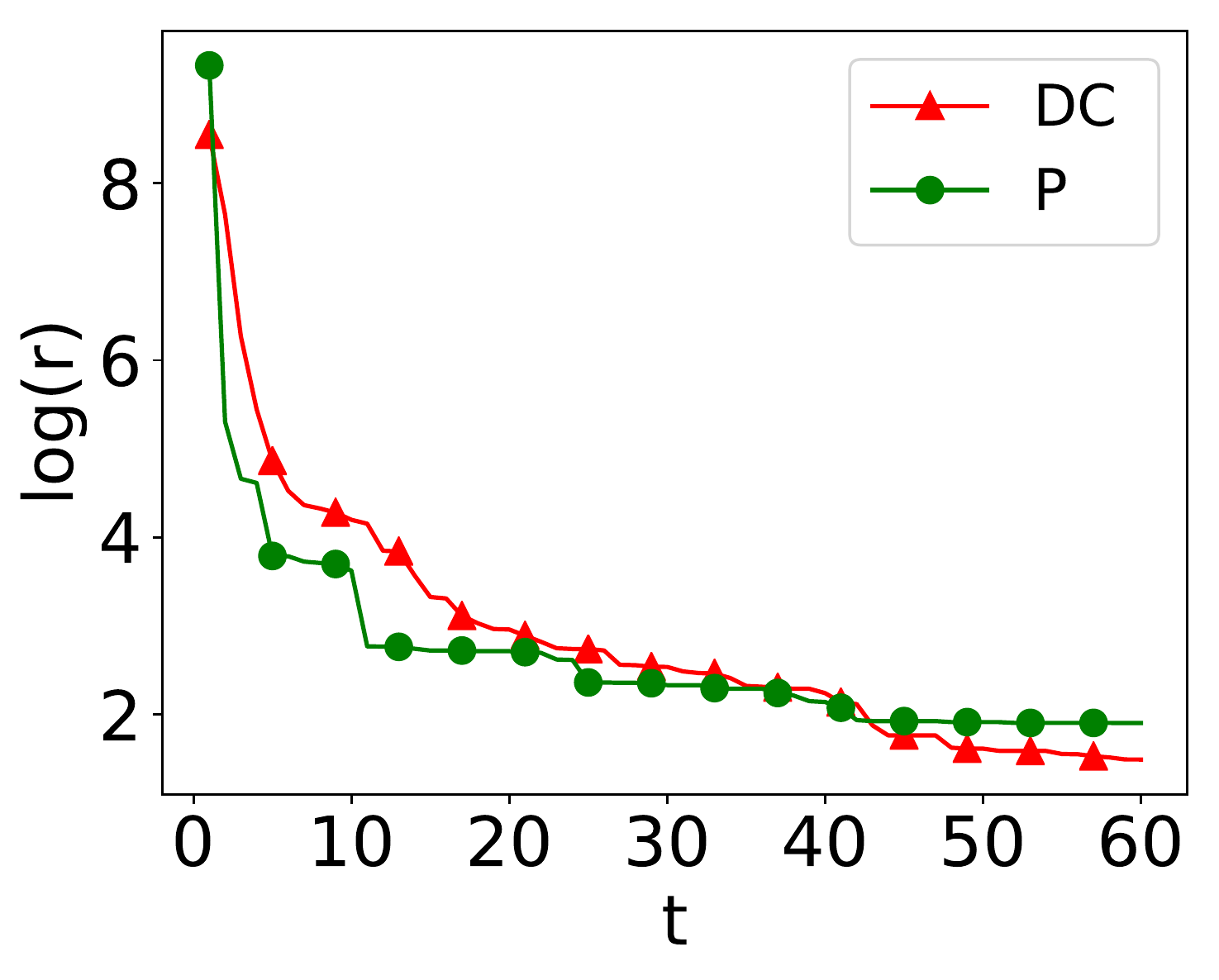}}	
	\subfigure[\scriptsize Industrial-2 (d=6)]{\includegraphics[width=0.26\textwidth]{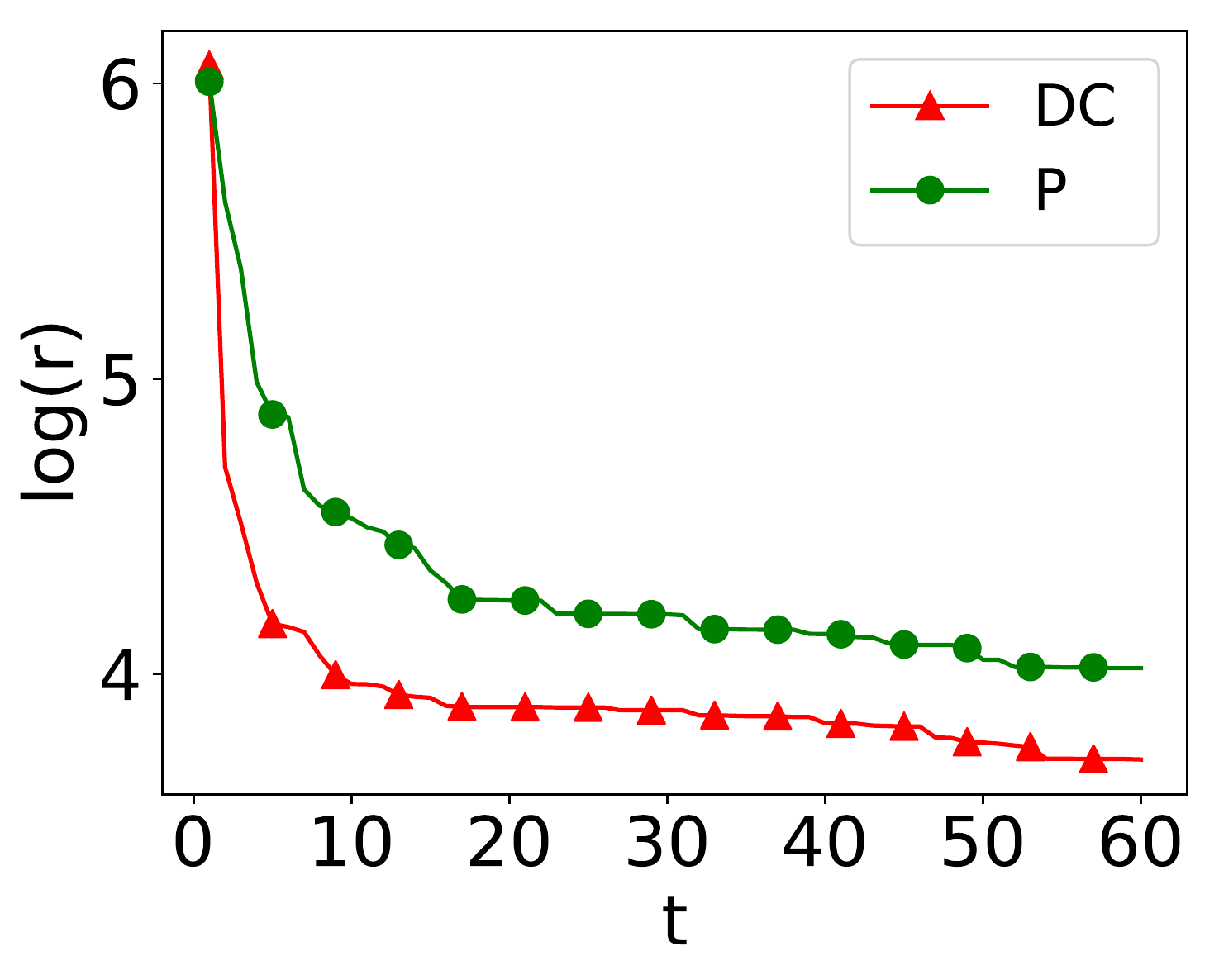}}	
	
	\caption{Regret plots comparing RNN-Opt-DC (DC) and RNN-Opt-P (P). The entries in the brackets denote values for $(\tau_1,\tau_2)$ for GMM-DF, and $\tau$ for Rosenbrock.\label{fig:rnn-opt-dc-vs-p}}
\end{figure}

We use the standard quadratic (disk) constraint used to evaluate constrained optimization approaches, i.e. $||\mathbf{x}||_2^2\leq \tau \times d$ (we use $\tau = \{0.5, 1.0, 2.0\}$) for Rosenbrock function. 
For GMM-DF, we generate random limit constraints on each dimension around the global optima, s.t. the optimal solution is still the same as the one without constraints, while the feasible search space varies randomly across functions.
Limits of the domain is $[\mathbf{x}_{opt}-\Delta \mathbf{x},\mathbf{x}_{opt}+\Delta \mathbf{x}]$, where $\Delta x^j$ ($j$-th component of $\Delta \mathbf{x}$) is sampled from $U(\tau_1,\tau_2)$ (we use $\tau_1=\{0.5,1.0,1.5\}$, $\tau_2=\{1.5,2.0,2.5\}$).
We also consider two instances of (anonymized) non-linear surrogate model for a real-world industrial process built by subject-matter experts with six controllable input parameters ($d=6$) as black-box functions, referred to as Industrial-1 and Industrial-2 in Fig. \ref{fig:rnn-opt-dc-vs-p}. 
This process imposes limit constraints on all six parameters guided by domain-knowledge. The ground-truth optimal value for these functions was obtained by querying the surrogate model ~200k times via grid search.
The regret results are averaged over runs assuming diverse environmental conditions.

RNN-Opt-DC and RNN-Opt-P are not guaranteed to propose feasible queries at all steps because of the soft constraints during training and/or inference. Therefore, despite training the optimizers for $T$ steps we unroll the RNNs up to a maximum of $5T$ steps and take the first $T$ proposed queries that are feasible, i.e. satisfy domain constraints. For functions where optimizer is not able to propose $T$ feasible queries in $5T$ steps, we replicate the regret corresponding to best solution for remaining steps. 
As shown in Fig. \ref{fig:rnn-opt-dc-vs-p}, we observe that \textbf{RNN-Opt with domain constraints, namely, \textit{RNN-Opt-DC} is able to effectively use explicit penalty feedback, and at least as good as RNN-Opt-P in all cases.} As expected, we also observe that the performance of both optimizers degrades with increasing values of $\tau$ or $\tau_2 - \tau_1$ as the search space to be explored by the optimizer increases.

\section{Conclusion and Future Work\label{sec:discussion}}
Learning optimization algorithms under the meta-learning paradigm is an area of active research. 
In this work, we have shown that using regret directly as a loss for training optimizers using recurrent neural networks is possible, and that it yields better optimizers than those obtained using observed-improvement based loss. 
We have proposed useful extensions of practical importance to optimization algorithms for black-box optimization that allow dealing with diverse range of function values and handle domain constraints more effectively.
One shortcoming of this approach is that a different optimizer needs to be trained for varying number of input parameters. 
In future, we plan to extend this work to train optimizers that can ingest input with varying and high number of parameters, e.g. by first proposing a change in a latent space and then estimating changes in actual input space as in \cite{rusu2018meta,wichrowska2017learned}.
Further, training optimizers for multi-objective optimization can be a useful extension.

\appendix

\section{Generating Diverse Non-Convex Synthetic Functions\label{apx:gmm-df}}
We generate synthetic non-convex continuous functions $f_g$ defined over $\Theta \subseteq \mathbb{R}^d$ via a Gaussian Mixture Model density function (GMM-DF, similar to \cite{zhou2017optimizing}):
\begin{equation}
	f_g(\mathbf{x}_t)= \sum_{i=1}^{N}\frac{c_i}{(2\pi)^{\frac{k}{2}}|\mathbf{\Sigma}_i|^{\frac{1}{2}}}\exp(-\frac{1}{2}(\mathbf{x}_t-\mathbf{\boldsymbol{\mu}}_i)^T\mathbf{\Sigma}^{-1}_i(\mathbf{x}_t-\boldsymbol{\mu}_i)).\label{eq:gmm}
\end{equation}
In this work, we used GMM-DF instead of Gaussian Processes used in \cite{chen2017learning} for ease of implementation and faster response time to queries:
\begin{figure}
	\subfigure{\includegraphics[width=0.3\textwidth,trim={2cm, 1cm, 1cm, 2cm},clip]{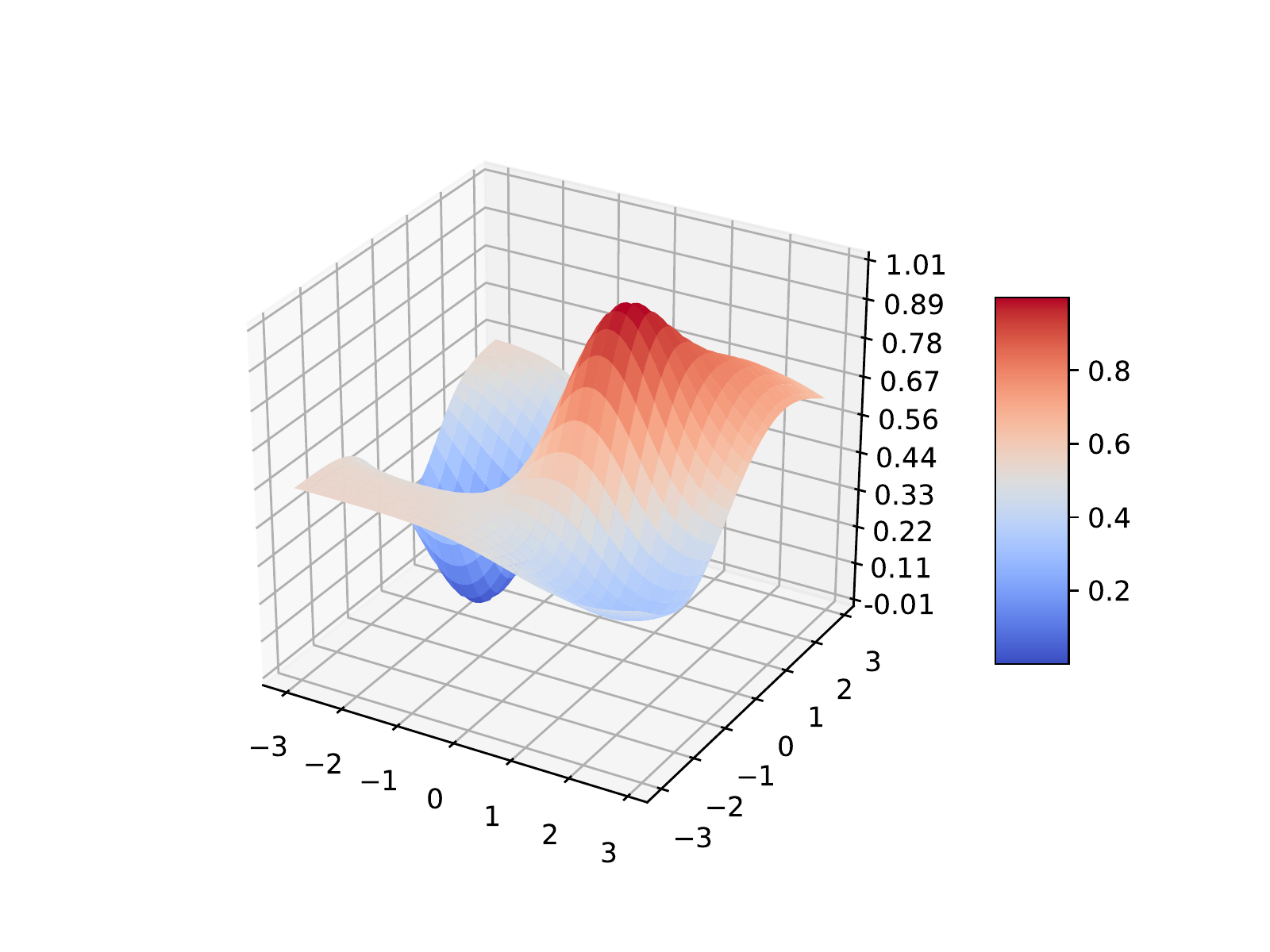}}
	\subfigure{\includegraphics[width=0.3\textwidth,trim={2cm, 1cm, 1cm, 2cm},clip]{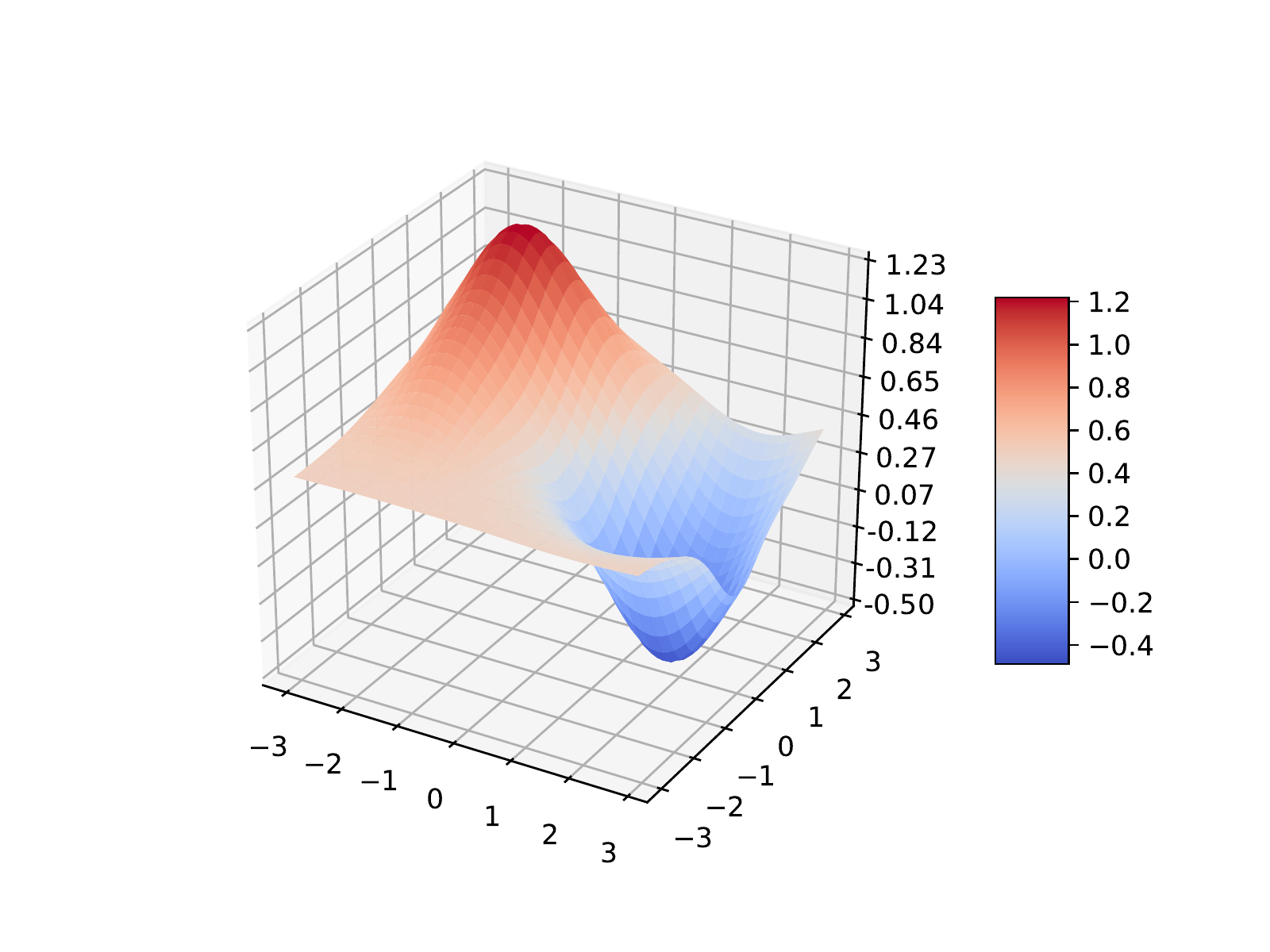}}
	\subfigure{\includegraphics[width=0.3\textwidth,trim={2cm, 1cm, 1cm, 2cm},clip]{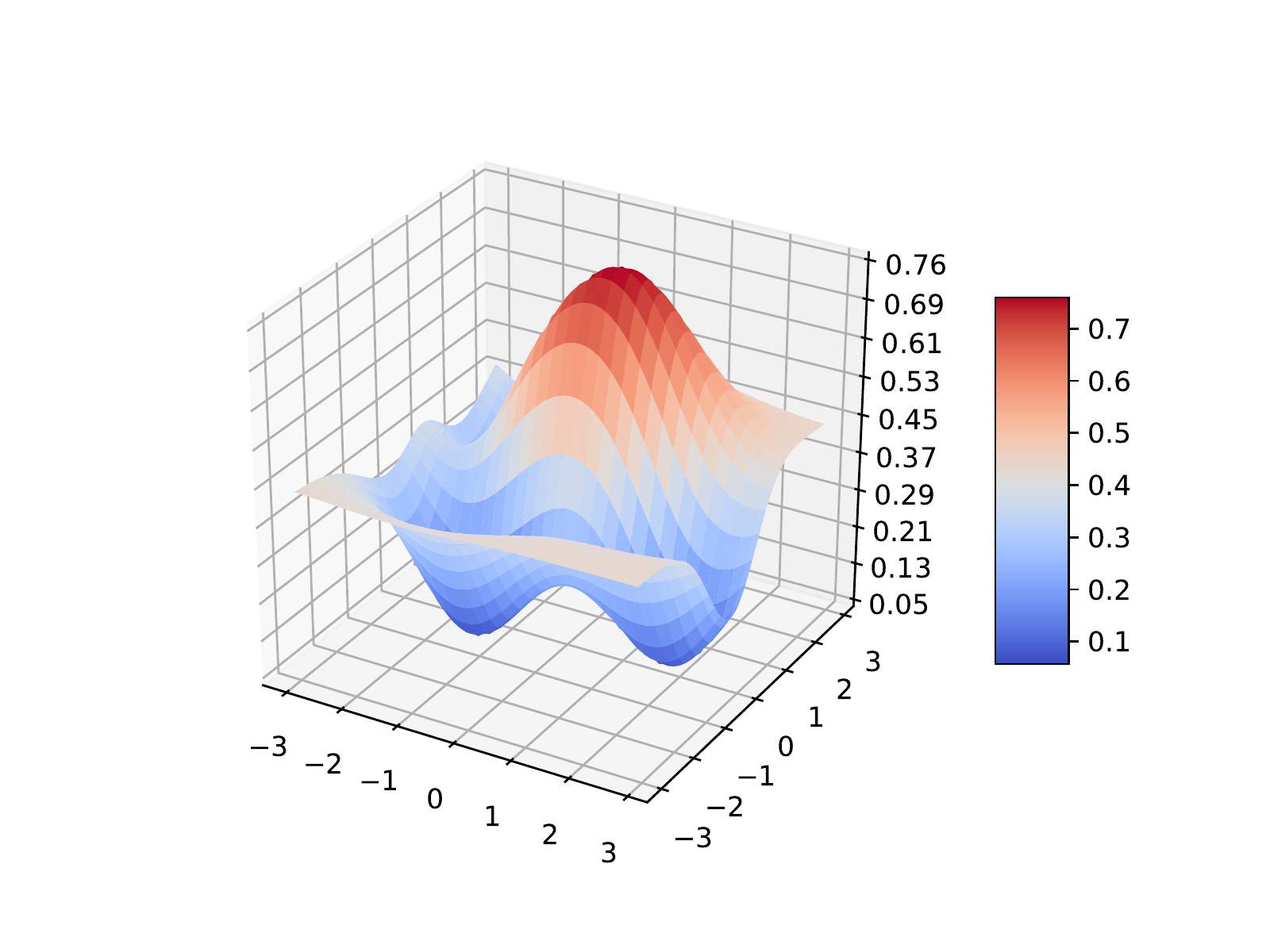}}
	\caption{Sample synthetic GMM density functions for $d=2$.}
	\label{fig:gmm-df}
\end{figure}
Functions obtained in this manner are often non-convex and have multiple local minima/maxima. 
Sample plots for functions obtained over 2-D input space are shown in Fig. \ref{fig:gmm-df}.
We use $c_i \sim \mathcal{N}(0,0.2)$, $\boldsymbol{\mu}_i \sim U(-2.0,2.0)$ and $\mathbf{\Sigma}_i \sim Truncated\mathcal{N}(0.9,0.9/5)$ for $d=2$, $\boldsymbol{\mu}_i \sim U(-3.0,3.0)$ and $\mathbf{\Sigma}_i \sim Truncated\mathcal{N}(3.0,3.0/5)$ for $d=6$  in our experiments (all covariance matrices are diagonal).

For any function $f_g$, we use an estimated value $\hat{y}_{opt} = \max_{i} f_g(\boldsymbol{\mu}_i)$ ($i=1,2,\dots,N$) instead of $y_{opt}$. This assumes that the global maximum of the function is at the mean of one of the $N$ Gaussian components. We validate this assumption by obtaining better estimates of the ground truth for $y_{opt}$ via grid search over randomly sampled 0.2M query points over the domain of $f_g$. For 10k randomly sampled GMM-DF functions, we obtained an average error of 0.03 with standard deviation of 0.02 in estimating $y_{opt}$, suggesting that the assumption is reasonable, and in practice, approximate values of $y_{opt}$ suffice to estimate the regret values for supervision.
However, in general, $y_{opt}$ can also be obtained using gradient descent on $f_g$.
\bibliographystyle{splncs04}

\end{document}